%% file: main.tex
\documentclass[10pt,twocolumn,letterpaper]{article}

\usepackage{iccv}              

\input{preamble}
\usepackage{graphicx}
\usepackage{amsmath}
\usepackage{amssymb}
\usepackage{bm}
\usepackage[export]{adjustbox}
\usepackage{dsfont}
\usepackage{pifont}
\usepackage{soul}
\usepackage{colortbl}
\usepackage{bbm}
\usepackage{wrapfig,lipsum,booktabs}
 
\usepackage[bottom]{footmisc}
\usepackage{float}
\usepackage{placeins}

\definecolor{iccvblue}{rgb}{0.21,0.49,0.74}
\usepackage[pagebackref,breaklinks,colorlinks,allcolors=iccvblue]{hyperref}

\def\paperID{9} 
\def\confName{ICCV}
\def\confYear{2025}

\title{Synthetic Captions for Open-Vocabulary Zero-Shot Segmentation}

\author{
\centerline{Tim Lebailly$^{1,2}$ \hspace{1cm} Vijay Veerabadran$^1$ \hspace{1cm} Satwik Kottur$^1$}\\
\centerline{Karl Ridgeway$^1$ \hspace{1cm} Michael Louis Iuzzolino$^1$} \\
\centerline{$^{1}$Meta \hspace{1cm} $^{2}$KU Leuven} \\ 
}

\begin{document}
\maketitle
\input{sec/def}

\input{sec/abstract}
\input{figures_tex/vlm_comparison}
\input{figures_tex/schema}

\input{sec/introduction}
\input{sec/related_works}
\input{sec/method}
\input{sec/experiments}
\input{sec/conclusion}
{
    \small
    \bibliographystyle{ieeenat_fullname}
    \bibliography{main}
}

\end{document}

%% file: sec/def.tex
\newcommand{\prompt}{\boldsymbol{p}}
\newcommand{\VLM}{f_{\texttt{VLM}}}

\newcommand{\im}{\boldsymbol{x}_v}
\newcommand{\txt}{\boldsymbol{x}_{t}}
\newcommand{\vglob}{\bar{\boldsymbol{z}}_{v}}
\newcommand{\tglob}{\bar{\boldsymbol{z}}_{t}}

\newcommand{\vglobb}{\bar{\boldsymbol{Z}}_{v}}
\newcommand{\tglobb}{\bar{\boldsymbol{Z}}_{t}}

\newcommand{\cent}{\boldsymbol{c}}
\newcommand{\q}{\boldsymbol{q}}
\newcommand\norm[1]{\left\lVert#1\right\rVert}

\newcommand{\NN}{\texttt{nn}}
\newcommand{\jump}{\texttt{jump}}
\newcommand{\cyclecond}{\mathcal{O}}

\newcommand{\vdense}{\boldsymbol{z}_{v}}
\newcommand{\tdense}{\boldsymbol{z}_{t}}
\newcommand{\pos}{\mathbf{E}}
\newcommand{\objects}{\mathbf{O}}
\newcommand{\vconcept}{\mathbf{c}_{v}}
\newcommand{\tconcept}{\mathbf{c}_{t}}

\newcommand{\lit}{\text{LiT}}
\newcommand{\litemoji}{\text{LiT} \raisebox{-0.15\height}{\includegraphics[width=2.5mm]{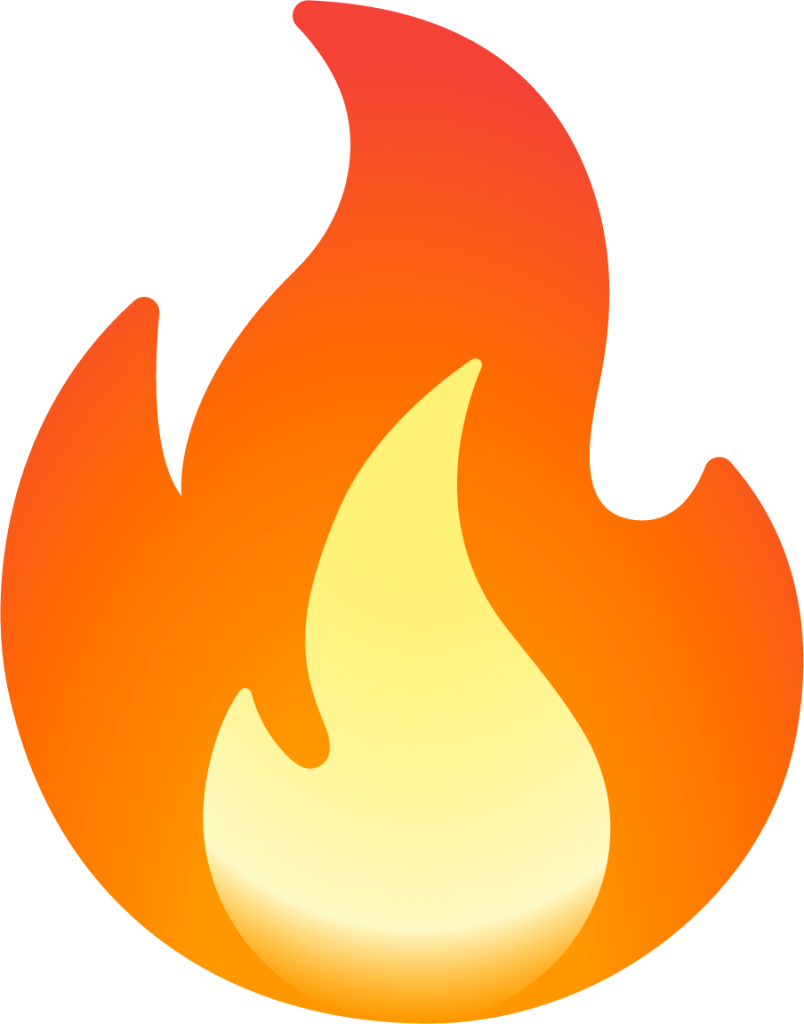}}\hspace{0.3mm}}

\newcommand{\vit}{\text{ViT}}
\newcommand{\croc}{\text{CrOC}}
\newcommand{\dino}{\text{DINO}}
\newcommand{\mae}{\text{MAE}}
\newcommand{\ibot}{\text{iBOT}}
\newcommand{\reco}{\text{ReCo}}
\newcommand{\viewco}{\text{ViewCo}}
\newcommand{\ovdiff}{\text{OVDiff}}
\newcommand{\groupvit}{\text{GroupViT}}
\newcommand{\zeroseg}{\text{ZeroSeg}}
\newcommand{\segclip}{\text{SegCLIP}}
\newcommand{\tcl}{\text{TCL}}
\newcommand{\clippy}{\text{CLIPpy}}
\newcommand{\ovsegmentor}{\text{OVSegmentor}}
\newcommand{\clipdiy}{\text{CLIP-DIY}}
\newcommand{\maskclip}{\text{MaskCLIP}}
\newcommand{\clipdinoiser}{\text{CLIP-DINOiser}}
\newcommand{\mname}{\text{SimZSS}}

\definecolor{light_cyan}{HTML}{d9f0ff}
\definecolor{light_green}{HTML}{72B54D}
\sethlcolor{light_cyan}

%% file: sec/abstract.tex
\begin{abstract}
Generative vision-language models (VLMs) exhibit strong high-level image understanding but lack spatially dense alignment between vision and language modalities, as our findings indicate. Orthogonal to advancements in generative VLMs, another line of research has focused on representation learning for vision-language alignment, targeting zero-shot inference for dense tasks like segmentation. In this work, we bridge these two directions by densely aligning images with synthetic descriptions generated by VLMs. Synthetic captions are inexpensive, scalable, and easy to generate, making them an excellent source of high-level semantic understanding for dense alignment methods. Empirically, our approach outperforms prior work on standard zero-shot open-vocabulary segmentation benchmarks/datasets, while also being more data-efficient. 
\end{abstract}

%% file: figures_tex/vlm_comparison.tex
\begin{figure}[b]
  \centering
  \includegraphics[width=\linewidth]{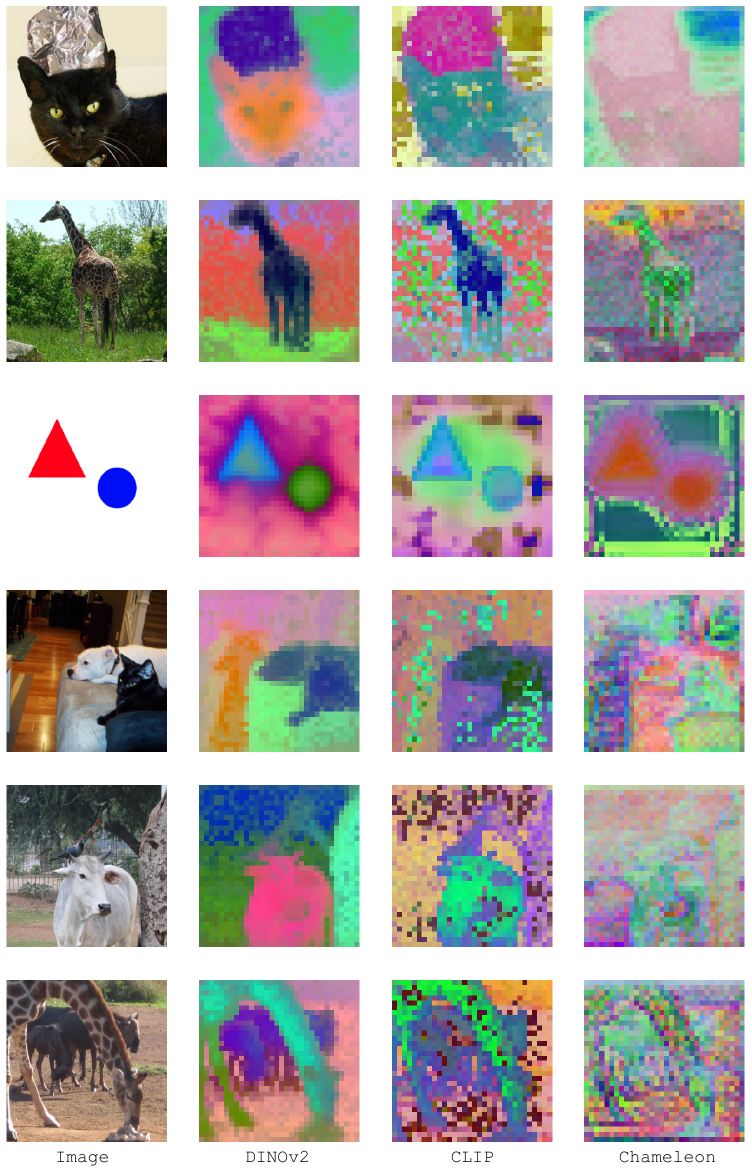}
  \caption{\textbf{Comparison of visual features with PCA projection.} Each $i$-th visual token (\ie local representation) $\vdense^i \in \mathbb{R}^{d_{v}}$ is mapped to $\mathbb{R}^{3}$ with a PCA projection and visualized in RGB space. The different visual encoders are (from left to right) DINOv2 \cite{oquab2023dinov2}, CLIP \cite{radford2021learning} and Chameleon \cite{chameleon}. The self-supervised pretraining from DINOv2 leads to a spatially coherent dense feature map where objects (or semantic entities) are represented as a unified color.}
  \label{fig:vlm_comparison}
\end{figure}

%% file: figures_tex/schema.tex
\begin{figure*}[t]
  \centering
  \includegraphics[width=\linewidth]{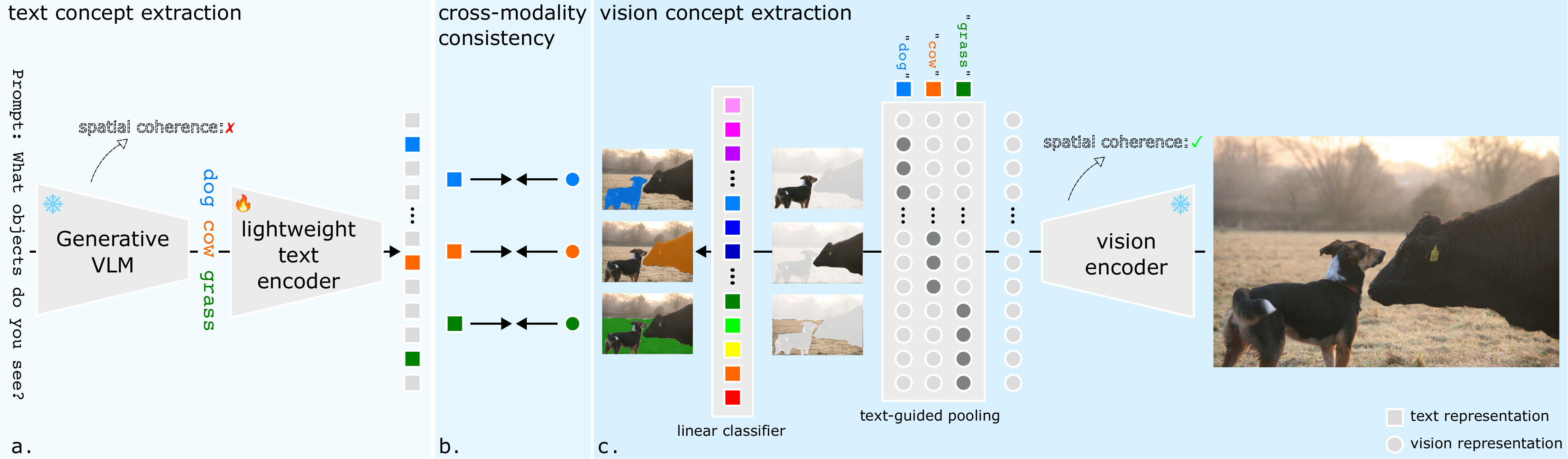}
  \caption{\textbf{Schematic overview of the method.} \textbf{(a.)} Given an image \(\im\) and a prompt \(\prompt\), a synthetic description of the image is generated \(\txt = \VLM(\im, \prompt)\). These synthetic descriptions provide a rich source of semantic information that can be used to identify the concept representations in both modalities. The concepts extracted from the synthetic description are then mapped to textual representations $\tconcept^{l}$ through a lightweight text encoder. \textbf{(c.)} These textual representations are used to pool the features of a spatially coherent vision encoder (DINOv2~\cite{oquab2023dinov2}) to obtain corresponding visual representations $\vconcept^{l}$. \textbf{(b.)} A cross-modal constraint is enforced between pairs of concepts in each modality. Note that the vision features of the VLM are not spatially coherent, while those from the dedicated vision encoder are (see \Cref{fig:vlm_comparison}). The image is also fed to the VLM but is omitted for simplicity. All models are frozen except for the text encoder.\protect\footnotemark}
  \label{fig:overview_sketch}
\end{figure*}

%% file: sec/introduction.tex
\section{Introduction}
\label{sec:intro}

Segmentation is a crucial task in computer vision, enabling the precise delineation of objects within an image. Traditional segmentation approaches \cite{DBLP:journals/corr/RonnebergerFB15, DBLP:journals/corr/HeGDG17} heavily rely on semantic segmentation annotations, which are both time and labor-intensive to gather. This reliance not only limits the applicability of these methods to domains where labels have been collected but also constrains the potential for broader generalization.
Recently, Segment Anything (SAM) \cite{segmentanything} has demonstrated remarkable performance through advanced engineering and the scaling of data and compute resources. While the resulting segmentations are impressive, this approach remains constrained by the need for human annotations, which is sub-optimal from a learning paradigm perspective.

In contrast, other emerging approaches aim to eliminate the dependency on human-annotated segmentation masks.
Most of these alternative methods leverage weak supervision through image-caption alignment pretraining \cite{radford2021learning,lit_beyer,jia2021scaling}. These image-caption pairs are typically scraped from the web without human review. The prevalent strategy involves globally aligning vision and language modalities, where a global image representation is matched with a global representation of the caption. However, this global alignment is sub-optimal as it entangles multiple entities present in the image-caption pair.
A more recent line of work seeks to densely align vision and language modalities, implicitly matching semantic entities in the caption to image sub-regions \cite{cha2023learning,simzss}. While conceptually elegant, this alignment is contingent on the specific image-caption pairs available. If the caption includes semantic entities not present in the image, or vice versa, it results in an ill-defined alignment objective.

Orthogonal to these developments, generative Vision-Language Models (VLMs) have gained significant popularity \cite{liu2023llava,chameleon,liu2024llavanext}. They can effectively describe images in an open-vocabulary setting and exhibit a strong high-level understanding of images. Although excellent at image-level tasks, our findings indicate that the vision and language representations in VLMs still lack spatially coherent features, as illustrated in the last two columns of \Cref{fig:vlm_comparison}. This deficiency, observed by heterogeneous colors in unique semantic entities (\eg an object), renders them inadequate for dense tasks like segmentation.

Our aim is to capitalize on the strengths of both lines of work, that is, 1) the high-level understanding of VLMs for generating synthetic image descriptions, and 2) the explicit dense alignment of certain vision-language methods. By synergizing these approaches, we eliminate the dependency on human annotated image-caption pairs, allowing for a more consistently defined alignment objective. Our main contributions are as follows:
\begin{itemize}[leftmargin=0.8cm]
    \item[(i)] We propose a dense-vision alignment method that aligns synthetic captions from a generative VLM with arbitrary sets of images, illustrated in \Cref{fig:overview_sketch}.
    \item[(ii)] We provide empirical evidence of the effectiveness of the proposed method by outperforming prior arts on standard open-vocabulary zero-shot segmentation benchmarks/datasets.
\end{itemize}

%% file: sec/related_works.tex
\footnotetext{The figure is adapted from \cite{simzss}.}

\section{Related Works}

\subsection{Vision Language Models (VLMs)}
\subsubsection{Contrastive Learning Based VLMs}

Recent advances in vision-language representation learning have been significantly influenced by CLIP~\cite{radford2021learning}, which established a foundational framework for contrastive learning using weakly-supervised image-caption pairs~\cite{radford2021learning,jia2021scaling,lit_beyer,zhai2023sigmoidlosslanguageimage}. This approach builds upon successful contrastive learning techniques developed in self-supervised learning (SSL)~\cite{chen2020simple,hjelm2018learning,he2020momentum}. The core mechanism operates by maximizing the similarity between corresponding image-caption pairs in a shared embedding space while minimizing the similarity between non-corresponding pairs through a contrastive objective.
Subsequent research has explored various directions to enhance this paradigm. ALIGN~\cite{jia2021scaling} demonstrated the scalability of this approach, while other works have focused on improving the fundamental alignment objective. 

One drawback of these methods is their reliance on negative samples in the contrastive loss computation, necessitating large batch sizes during training. To address this constraint, \citet{zhai2023sigmoidlosslanguageimage} proposed SigLIP, which eliminates batch size dependency by adopting an NCE objective~\cite{infonce} in place of the InfoNCE loss~\cite{nce_loss}. Complementary advances include the work of \citet{lit_beyer}, who achieved sped-up training through the use of frozen vision encoders, and \citet{lavoie2024modelingcaptiondiversitycontrastive}, who enhanced model performance by incorporating caption diversity modeling into the image encoding process. It is worth noting that the majority of existing contrastive vision-language methods, including those discussed above, operate by globally aligning image-level representations with sentence-level representations.

\subsubsection{Generative VLMs}
Unlike to contrastive learning-based approaches, generative methods aim to directly produce vision or language data~\cite{tsimpoukelli2021multimodal,yu2022coca,liu2023llava,liu2024llavanext,chameleon,DBLP:journals/corr/abs-2201-12086,li2023blip2bootstrappinglanguageimagepretraining,alayrac2022flamingovisuallanguagemodel,yu2023scalingautoregressivemultimodalmodels}. Training such vision-language models (VLMs) typically requires substantial computational resources, which can be prohibitive for many researchers.
To address this challenge, a popular research direction has been to leverage the strengths of pre-trained models. Specifically, vision encoders with features learned through vision-language contrastive pretraining are often combined with open-source large language models (LLMs) such as LLaMA~\cite{touvron2023llamaopenefficientfoundation,touvron2023llama2openfoundation,dubey2024llama3herdmodels} or Mistral-7B~\cite{jiang2023mistral7b}. The modalities are then integrated through techniques like cross-attention~\cite{alayrac2022flamingovisuallanguagemodel} or adapter layers on the vision side~\cite{DBLP:journals/corr/abs-2201-12086,li2023blip2bootstrappinglanguageimagepretraining,liu2023llava,liu2024llavanext}.
More recently, alternative approaches have emerged that tokenize visual data and utilize a unified autoregressive objective to predict the next token, with special delimiter tokens used to delineate the two modalities~\cite{yu2023scalingautoregressivemultimodalmodels,chameleon}. These methods have the advantage of being able to generate both vision and language outputs.

\subsection{Open-Vocabulary Zero-Shot Segmentation}

Typical computer vision models operate under the closed-vocabulary assumption, where the object categories to classify, detect, or segment are predetermined during training. This limits generalization and requires extensive labeling. Open-vocabulary learning aims to address these limitations. Here,  contrastive learning based VLMs are often repurposed~\cite{radford2021learning,zhai2023sigmoidlosslanguageimage}. In fact, open-vocabulary zero-shot classification naturally arises from such VLMs due to the vision-language alignment. However, this alignment is only global in nature, meaning that the local pixels are not densely aligned with the corresponding text describing those pixels, making open-vocabulary zero-shot segmentation impossible.

To overcome these issues, a dedicated branch of research has focused on learning a dense vision-language alignment, without relying on any segmentation annotations~\cite{mukhoti2023open,xu2022groupvit,cha2023learning,wysoczanska2024clip,wysoczańska2024clipdinoiser,rewatbowornwong2023zero,simzss}. \citet{xu2022groupvit} leverage hierarchical grouping of visual tokens based on learnable query tokens similarity. \citet{mukhoti2023open} learn a special pooling layer where the patch-level representations in the vision encoder are weighted based on their similarity with the caption, bootstrapping the initial alignment. \citet{cha2023learning} learn to align only the foreground object with the caption through learnable masking. \cite{wysoczanska2024clip,wysoczańska2024clipdinoiser,rewatbowornwong2023zero} leverage self-supervised learning (SSL) trained vision features~\cite{caron2021emerging} as well as a vision encoder that has been contrastively aligned with language~\cite{radford2021learning}. This combination aims to capture both the spatially coherent visual representations from SSL pretraining and the cross-modal alignment from contrastive vision-language learning. Finally, the state-of-the-art method SimZSS~\cite{simzss} proposes a simple framework for dense vision-language alignment. Instead of relying on two separate vision encoders, SimZSS disentangles the visual representation learning process from the dense alignment process by leveraging a frozen SSL-trained vision encoder~\cite{oquab2023dinov2}, and densely aligning a text-encoder similar to the approach used in LiT~\cite{lit_beyer}. Orthogonally, they propose an object-level alignment loss that locally aligns query concepts in captions with their corresponding visual counterpart. It is important to note that the performance is inherently dependent on the quality of the underlying image-caption pairs. The assumption that all textual concepts in the caption are also present in the image may not always hold true.

%% file: sec/method.tex
\section{Method}
In this section, we start by defining the notations and nomenclature from \cite{simzss} in \Cref{sec:notations} and the discuss the shortcomings of generative VLMs in \Cref{sec:shortcomings-vlms}. This foundation allows us to clearly articulate the process of dense alignment using weak supervision with synthetic captions in \Cref{sec:dense-vision-language-alignment}.

\subsection{Notations \& Nomenclature}
\label{sec:notations}
Transformers encode input signals as sequences of tokens, resulting in a \textbf{dense representation}. For vision transformers, this is a tensor \(\vdense \in \mathbb{R}^{n_{v} \times d_{v}}\), where \(d_{v}\) is the dimension of the representation space and \(n_{v}\) is the number of image patches. For text, the dense representation \(\tdense \in \mathbb{R}^{n_{t} \times d_{t}}\) consists of \(d_{t}\)-dimensional tokens. Indexing a dense representation with a sequence index \(i\) yields a \textbf{local representation}: \(\tdense^{i} \in \mathbb{R}^{d_{t}}\) for text data and \(\vdense^{i} \in \mathbb{R}^{d_{v}}\) for visual data. By aggregating local representations linked to the \(l^{th}\) semantic concept, we derive a \textbf{concept representation}: \(\vconcept^{l} \in \mathbb{R}^{d_{v}}\) for visual data and \(\tconcept^{l} \in \mathbb{R}^{d_{t}}\) for text data. The \textbf{global representation} \(\vglob \in \mathbb{R}^{d_{v}}\) of an entire input signal uses specialized tokens. For visual data, \(\vglob \in \mathbb{R}^{d_{v}}\) is the \texttt{[CLS]} token, while \(\tglob \in \mathbb{R}^{d_{t}}\) is the \texttt{[EOS]} token for text data. We assume $d_{v}$ is equal to $d_{t}$ without loss of generality (a projection layer can be added if $d_{v} \neq d_{t}$).

\subsection{Shortcomings of VLMs}
\label{sec:shortcomings-vlms}

While generative visual-language models (VLMs) have made impressive strides in high-level image understanding, they still fall short of achieving dense vision-language alignment, an essential component for local-level semantic understanding. True dense alignment requires two specific conditions.

\vspace{0.2cm}
\begin{itemize}[leftmargin=0.8cm]
    \item[(i)] Visual patch representations that correspond to the same object must exhibit similarity, while representations of distinct objects should exhibit dissimilarity.
    \item[(ii)] Visual representations must be aligned with the corresponding textual representations of each region. 
\end{itemize}
\vspace{0.2cm}

Observing a principal component analysis (PCA) of local vision features in \Cref{fig:vlm_comparison} makes it clear that this first condition is not satisfied in models like CLIP \cite{radford2021learning}. The contrastive pretraining used in CLIP, while effective for global image-to-text matching, fails to produce spatially coherent feature maps. Therefore, any VLM reliant on CLIP-like representations will inherently lack dense vision-language alignment. Surprisingly, even recent generative VLMs, capable of generating coherent images, exhibit this limitation (see Chameleon \cite{chameleon} in \Cref{fig:vlm_comparison}).

Nonetheless, these generative models possess a powerful, high-level grasp of image semantics that could be used to facilitate the learning of such dense alignment. We focus on integrating synthetic captions with SimZSS \cite{simzss}, the current state-of-the-art for dense vision-language alignment.

\subsection{Dense Vision-Language Alignment}
\label{sec:dense-vision-language-alignment}
Applying a dense cross-modal constraint to enforce consistency between pixels corresponding to a semantic entity and its textual description is not a straightforward task. Ideally, we would like to impose a consistency loss between the two modalities, but this essentially boils down to a set matching problem, illustrated in part (b.) of \Cref{fig:overview_sketch}). For example, the textual concept of \eg \textit{cow} should be matched and aligned with the visual concept of \textit{cow}. However, when dealing with noisy image-caption pairs, it is challenging to assume that there always exists a matchable element in the other set. For example, the word \textit{cow} might not be present in the caption but present in the image. To alleviate this issue, we leverage a Vision Language Model (VLM) to generate synthetic image captions, which produces much less noisy text descriptions.

Furthermore, given an image, it is not trivial to determine the set of vision concepts. Fortunately, thanks to the VLM, we can utilize text concepts to pool the dense vision features into a set of corresponding vision concepts. This approach enables us to effectively apply a dense cross-modal constraint, enforcing consistency between the visual and synthetic textual modalities as in \cite{simzss}.

\subsubsection{Concept Identification with Generative VLM}
\label{subsubsec:synthetic-captions}

Identifying the concepts present in an image is a complex and non-trivial task, which can be significantly aided by leveraging the language modality. Language provides a well-defined structure and can be efficiently analyzed using tools such as part-of-speech taggers (POS). We rely on a generative Vision-Language Model (VLM) to produce synthetic descriptions for images, which not only enables the identification of concepts in cases where no captions are available, but also could lead to better image descriptions as web-crawled or human-annotated captions are often incomplete, non-exhaustive, and occasionally irrelevant to the image content. Given an image \(\im\), a prompt \(\prompt\), and a generative VLM \(\VLM\), we can generate synthetic captions \(\txt\) (i.e., descriptions) of the image through the mapping \(\txt = \VLM(\im, \prompt)\). These descriptions, in turn, provide a rich source of semantic information that can be used to identify the concept representations in both modalities. 

The global and dense representations of both modalities are obtained through dedicated encoders, $(\tglob, \tdense)=f_t(\txt)$ and $(\vglob, \vdense)=f_v(\im)$. 

To identify key textual concepts, noun phrases (NPs) are extracted using part-of-speech (POS) tagging. Each NP corresponds to a span of indices within the dense textual representation $\tdense$. A concept representation $\tconcept^{l}$ is then derived by pooling over the dense representation at these indices as expressed in the following equation

\begin{equation}
    \tconcept^{l} = \frac{1}{|\mathcal{S}_{l}|} \sum_{i \in \mathcal{S}_{l}} \tdense^{i}
\end{equation}

where \( \mathcal{S}_{l} \) represents the set of indices corresponding to the \( l \)-th concept. 

The current vision-language alignment (between $f_t$ and $f_v$) can be bootstrapped to derive visual concept representation based on the textual concept representations. The former ($\vconcept^{l}$) are obtained by soft pooling the visual dense representations $\vdense$ proportionally to the similarity with a given textual concept representation $\tconcept^{l}$:
\begin{equation}
    \label{eq:pooling}
    \vconcept^l = \vdense^{\top} \texttt{softmax} \left( \frac{\vdense \mathbf{c}_{t}^{l}}{\tau} \right)
\end{equation}
where $\tau$ modulates the sharpness of the pooling distribution.

\input{tables_tex/segmentation_without_bkg}
\input{figures_tex/laion_scaling_tex}

\subsubsection{Alignment Objective}
\label{subsubsec:objective}

We adopt the alignment objective introduced by \cite{simzss}, which comprises two key components: a global-level alignment loss, $\mathcal{L}_{g}$, and a concept-level alignment loss, $\mathcal{L}_{l}$. The global-level loss, $\mathcal{L}_{g}$, is a contrastive loss inspired by the CLIP \cite{radford2021learning}, which encourages alignment between global representations of image-text pairs. The concept-level alignment loss, $\mathcal{L}_{l}$, deviates from traditional instance-level approaches, as individual concepts can occur across multiple captions or within a single caption, rendering instance-based methods unsuitable. Instead, we leverage the discrete nature of these concepts to utilize a cross-entropy loss. Here, each visual concept is intended to be classified as the label $c \in [C]$\footnote{No ground truth labels are used and the term \textit{label} is used as an identifier for a unique concept. As such, these labels simply define a ordering of all concepts \eg an alphabetical ordering.} of the corresponding textual concept that was used for the soft pooling in \Cref{eq:pooling}. Concretely, each visual concept $\vconcept^l$ is projected onto logits via a linear layer $\mathbf{h}$, which produces a probability distribution $\mathbf{p}$ over possible concepts.

\begin{equation}
    \label{eq:concepts_predictions}
    \mathbf{p} = \texttt{softmax}  \left( \vconcept^l \mathbf{h}^\top \right)
\end{equation}

We define $\mathbf{q}$ as the Kronecker delta distribution which is 1 for the concept with label $c$. The concept-level loss $\mathcal{L}_{l}$ is defined as the cross-entropy between $\mathbf{q}$ and $\mathbf{p}$, summed over all concepts in the batch. The total alignment objective is a weighted sum of both terms
\begin{equation}
    \label{eq:total_loss}
    \mathcal{L}_{\text{tot}} = \mathcal{L}_{g} + \lambda  \mathcal{L}_{l}
\end{equation}
where $\lambda$ is a weighting parameter.

\subsubsection{Training procedure}
\label{sec:training-procedure}

The lightweight text encoder in our model employs the same architecture as utilized in CLIP, but is randomly initialized and is the sole trainable network in the entire pipeline. All other networks are pretrained and remain frozen throughout the training. Specifically, the vision encoder is a frozen DINOv2 model~\cite{oquab2023dinov2}, chosen due to its representations which fulfill the first criterion outlined in \Cref{sec:shortcomings-vlms}. The VLM is chosen as a frozen LLaVA-NeXT-Mistral-7B model~\cite{liu2024llavanext}, selected for its extensive adoption, although many other VLMs would also be suitable here. 

Leveraging synthetic captions enables us to align the model with any image dataset, even those lacking explicit image captions. To facilitate comparisons with baseline methods (i.e., models that do not utilize synthetic captions), we use two image-caption pair datasets: COCO-Captions~\cite{chen2015microsoft} and a filtered subset of LAION-400M~\cite{schuhmann2021laion}, containing 10 million samples (denoted LAION-10M). Our alignment process begins by training on these datasets, ignoring their associated captions. Subsequently, since image-caption pairs are not strictly required, we further propose finetuning our model on the combined set of downstream datasets to improve alignment with the target data distributions. These downstream datasets include Pascal VOC~\citep{pascal-voc-2012}, Pascal Context~\citep{mottaghi2014role}, COCO~\citep{caesar2018coco}, Cityscapes~\citep{cordts2016cityscapes}, and ADE20K~\citep{zhou2017scene}.

The training is performed with the Adam optimizer~\cite{kingma2014adam} and a batchsize of 16384. The models are trained for 6 epochs on COCO-Captions and 1 epoch on LAION-10M.
\input{tables_tex/segmentation_whole_image}

%% file: tables_tex/segmentation_without_bkg.tex
\begin{table*}[t]
\centering
\caption{
\textbf{Zero-shot foreground segmentation.} $\dagger$ denotes our own reproduction, otherwise, the results are taken from \cite{simzss}.
}

\footnotesize
\resizebox{1\textwidth}{!}{
\begin{tabular}{l c c c c c c c c c c c c c c c c}
\toprule
\textbf{Method} & \raisebox{-0.15\height}{\includegraphics[width=2.5mm]{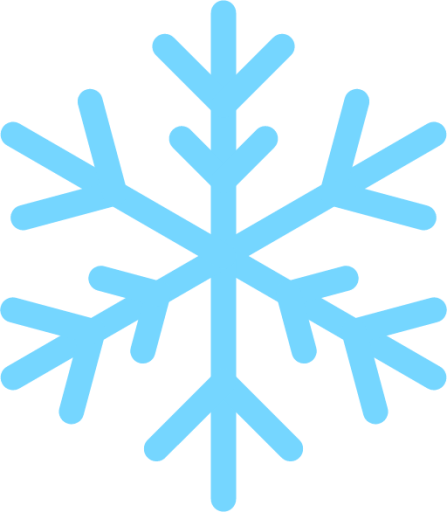}}\hspace{0.3mm} \textbf{Params} &  \raisebox{-0.15\height}{\includegraphics[width=2.5mm]{figures/learnable.png}}\hspace{0.3mm} \textbf{Params}  && \textbf{Pascal VOC} && \textbf{Pascal Context} &&\textbf{ COCO-Stuff} && \textbf{Cityscapes} && \textbf{ADE20K} && \textbf{Avg.}\\
\midrule
\textcolor{gray}{Miscellaneous} \\
$\reco$~\citep{shin2022reco}             & 313M & 0 && 57.7 && 22.3 && 14.8 && 21.1 && 11.2 && 25.4\\
$\groupvit$~\citep{xu2022groupvit}       & 0 & 55M  && 79.7 && 23.4 && 15.3 && 11.1 && 9.2 && 27.7\\
$\tcl$~\citep{cha2023learning}           & 156M & 21M  && 77.5 &&  30.3 &&  19.6 && 23.1 && 14.9 && 33.1\\
$\maskclip$~\citep{Dong_2023_CVPR}       & 291M & 0 && 74.9 && 26.4 && 16.4 && 12.6 && 9.8 && 28.0 \\
$\ovdiff$~\citep{karazija2023diffusion}  & 1,226M & 0  && 81.7 && 33.7 && - && - && 14.9 && -\\
$\clipdinoiser$~\citep{wysoczańska2024clipdinoiser}  & - & - && 80.9 && 35.9 && 24.6  && 31.7 && 20.0 && 38.6\\
LiT~\citep{lit_beyer} (ViT-B, LAION-400M)   & 94M & 63M && 80.5 && 31.8 && 23.3 && 24.7 && 18.7 && 35.8\\ 
LiT~\citep{lit_beyer} (ViT-B, COCO Captions)  & 94M & 63M  && 86.1 && 35.5 && 25.6 && 25.8 && 18.1 && 38.2\\ 
$\mname$~\cite{simzss} (ViT-B, LAION-400M) & 94M & 63M  && 85.1 && 34.2 && 24.9 && 27.8 && 19.6 && 38.3\\
$\mname$~\cite{simzss} (ViT-B, COCO Captions) & 94M & 63M  && 90.3 && 43.1 && 29.0 && 33.0 && 21.8 && 43.4\\
\arrayrulecolor{black!15}\midrule[0.25pt]

\textcolor{gray}{LAION-400M Subset (10M)} \\
Baseline (SimZSS$^{\dagger}$, ViT-B) & 94M & 63M && $80.8\pm0.58$ && $30.9\pm0.21$ && $22.9\pm0.24$ && $24.0\pm0.75$ && $19.4\pm0.27$ && $35.6\pm0.20$ \\ 
Baseline w/ synthetic captions & 94M & 63M  && $88.3\pm0.63$ && $40.3\pm0.62$ && $27.8\pm0.42$ && $30.2\pm0.17$ && $24.4\pm0.32$ && $42.2\pm0.35$ \\
\rowcolor{light_cyan} Baseline w/ synthetic captions + ft & 94M & 63M  && $88.8\pm0.46$ && $41.8\pm0.54$ && $28.7\pm0.49$ && $34.2\pm1.27$ && $24.4\pm0.63$ && $43.6\pm0.29$ \\

\arrayrulecolor{black!15}\midrule[0.25pt]
\textcolor{gray}{COCO Captions} \\

Baseline (SimZSS$^{\dagger}$, ViT-B) & 94M & 63M  && $89.6\pm0.41$ && $39.5\pm0.20$ && $27.3\pm0.22$ && $28.1\pm0.50$ && $19.6\pm0.18$ && $40.8\pm0.19$ \\
Baseline w/ synthetic captions & 94M & 63M  && $87.2\pm1.27$ && $42.5\pm0.18$ && $28.8\pm0.35$ && $36.6\pm0.35$ && $21.4\pm0.60$ && $43.3\pm0.24$ \\ 
\rowcolor{light_cyan} Baseline w/ synthetic captions + ft & 94M & 63M  && $89.2\pm0.51$ && $42.6\pm0.96$ && $29.0\pm0.32$ && $35.9\pm1.40$ && $23.6\pm0.09$ && $44.1\pm0.42$ \\ 

\end{tabular}
\label{table:zeroshot_without_bkg}
}
\end{table*}

%% file: figures_tex/laion_scaling_tex.tex
\begin{figure*}[t]
  \centering
  \includegraphics[width=\linewidth]{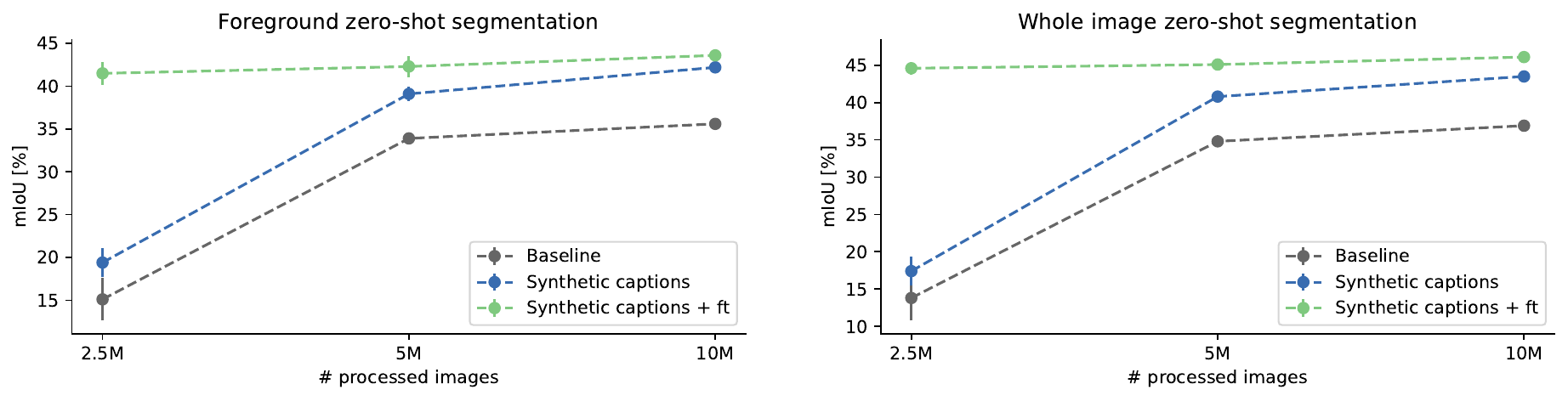}
  \caption{\textbf{Zero-shot segmentation performance as a function of the pretraining number of images.} Each plot compares 3 different settings: 1) the baseline (image-caption pairs), 2) using synthetic captions and 3) using synthetic captions plus an additional finetuning step.}
  \label{fig:laion_scaling}
\end{figure*}

%% file: tables_tex/segmentation_whole_image.tex
\begin{table*}[tp]
\centering
\caption{\textbf{Zero-shot whole-image segmentation.} $\dagger$ denotes our own reproduction, otherwise, the results are taken from \cite{simzss}.}

\footnotesize
\resizebox{\textwidth}{!}{
\begin{tabular}{l c c c c c c c c c c c c c}
\toprule
\textbf{Method} & \raisebox{-0.15\height}{\includegraphics[width=2.5mm]{figures/frozen.png}}\hspace{0.3mm} \textbf{Params} &  \raisebox{-0.15\height}{\includegraphics[width=2.5mm]{figures/learnable.png}}\hspace{0.3mm} \textbf{Params}  && \textbf{Pascal Context} && \textbf{COCO-Objec}t && \textbf{Pascal VOC} && \textbf{Avg.}\\
\midrule
\textcolor{gray}{Miscellaneous} \\
$\reco$~\citep{shin2022reco}                  & 313M & 0 && 19.9 && 15.7 && 25.1 && 20.2\\
$\ovdiff$~\citep{karazija2023diffusion}       & 1,226M & 0  && 30.1 && 34.8 && 67.1 && 44.0\\
$\groupvit$~\citep{xu2022groupvit}            & 0 & 55M && 18.7 && 27.5 && 50.4 && 32.2\\
$\zeroseg$~\citep{chen2023exploring}          & - & -  && 21.8 && 22.1 && 42.9 && 28.9\\
$\segclip$~\citep{luo2023segclip}             & - & - && 24.7 && 26.5 && 52.6 && 34.6\\
$\tcl$~\citep{cha2023learning}                & 156M & 21M && 24.3 && 30.4 && 51.2 && 35.3\\
$\clippy$~\citep{ranasinghe2023perceptual}    & - & -  && - && 32.0 && 52.2 && -\\
$\ovsegmentor$~\citep{xu2023learning}         & - & - && 20.4 && 25.1 && 53.8 && 33.1\\
$\clipdiy$~\citep{wysoczanska2024clip}        & - & - && 19.7 && 31.0 && 59.9 && 36.9\\
$\maskclip$~\citep{Dong_2023_CVPR}            & 291M & 0  && 23.6 && 20.6 && 38.8 && 27.7\\
$\clipdinoiser$~\citep{wysoczańska2024clipdinoiser}  & - & - && 32.4 && 34.8 && 62.1 && 43.1\\
LiT~\citep{lit_beyer} (ViT-B, LAION-400M) & 94M & 63M && 29.6 && 38.3 && 48.1 && 38.7\\
LiT~\citep{lit_beyer} (ViT-B, COCO Captions)  & 94M & 63M  && 31.5 && 39.5 && 51.4 && 40.8\\
$\mname$~\cite{simzss} (ViT-B, LAION-400M) & 94M & 63M && 31.1 && 38.1 && 48.6 && 39.3\\
$\mname$~\cite{simzss} (ViT-B, COCO Captions) & 94M & 63M  && 37.2 && 43.5 && 58.4 && 46.4\\

\arrayrulecolor{black!15}\midrule[0.25pt]
\textcolor{gray}{LAION-400M Subset (10M)} \\

Baseline (SimZSS$^{\dagger}$, ViT-B) & 94M & 63M && $29.1\pm0.12$ && $35.6\pm0.21$ && $45.9\pm0.13$ && $36.9\pm0.07$ \\
Baseline w/ synthetic captions & 94M & 63M && $35.4\pm0.55$ && $40.8\pm0.43$ && $54.2\pm1.37$ && $43.5\pm0.77$ \\ 
\rowcolor{light_cyan} Baseline w/ synthetic captions + ft & 94M & 63M  && $37.1\pm0.47$ && $41.7\pm0.27$ && $59.4\pm1.26$ && $46.1\pm0.56$ \\

\arrayrulecolor{black!15}\midrule[0.25pt]
\textcolor{gray}{COCO Captions} \\
Baseline (SimZSS$^{\dagger}$, ViT-B) & 94M & 63M && $34.1\pm0.16$ && $41.4\pm0.21$ && $54.0\pm0.26$ && $43.2\pm0.12$ \\ 
Baseline w/ synthetic captions & 94M & 63M && $37.1\pm0.25$ && $42.8\pm0.49$ && $61.3\pm1.94$ && $47.1\pm0.82$ \\ 
\rowcolor{light_cyan} Baseline w/ synthetic captions + ft & 94M & 63M  &&  $37.4\pm0.97$ && $42.6\pm0.29$ && $60.6\pm1.19$ && $46.9\pm0.72$ \\

\end{tabular}
\label{table:zeroshot_with_bkg}
}
\end{table*}%

%% file: sec/experiments.tex
\section{Experiments}
\label{sec:experiments}
In this section, we evaluate our method against prior arts and the state-of-the-art baseline that does not utilize synthetic captions on standard open-vocabulary zero-shot segmentation benchmarks (\cref{subsec:openvoc}, \cref{subsec:openvoc-wholeimage}). We also conduct an ablation study on the only hyperparameter introduced in our method, specifically the prompt used with the VLM (\cref{subsec:ablations}). Finally, we discuss noteworthy empirical observations (\cref{subsec:empirical-observations}).

\input{tables_tex/ablation_prompt}

\subsection{Foreground Open-Vocabulary Zero-Shot Segmentation}
\label{subsec:openvoc}
We evaluate the proposed method with a pixel-level zero-shot segmentation task, where the model depends entirely on textual class descriptions for pixel classification. Accurately representing the \texttt{background} class is challenging because captions often describe the background with terms like ``grass'' or ``floor'' rather than explicitly calling it ``background''. Therefore, consistent with previous research~\citep{wysoczańska2024clipdinoiser,cha2023learning,simzss}, we perform evaluations both without (\cref{table:zeroshot_without_bkg}) (foreground segmentation) and with (\cref{table:zeroshot_with_bkg}) the \texttt{background} class (whole image segmentation). 

We adopt the unified evaluation protocol introduced by TCL \cite{cha2023learning}, without applying any post-processing (e.g., DenseCRF \cite{densecrf}, PAMR~\citep{araslanov2020single}), to provide a better assessment of the intrinsic dense vision-language alignment. We resize images so that the shorter side measures 448 pixels, and perform inference with a sliding window of $448 \times 448$ and a stride of 224 pixels using the MMSegmentation~\citep{mmseg2020} implementation provided by \cite{cha2023learning}. Each class name is contextualized using ImageNet templates~\cite{radford2021learning} and fed to the text encoder to obtain its textual representation. Predictions are generated by projecting patch representations onto class names, followed by bilinear interpolation to up-sample these predictions to the original image dimensions. We report the mean Intersection over Union (mIoU) scores across five standard benchmarks: Pascal VOC~\citep{pascal-voc-2012}, Pascal Context~\citep{mottaghi2014role}, COCO-Stuff~\citep{caesar2018coco}, Cityscapes~\citep{cordts2016cityscapes}, and ADE20K~\citep{zhou2017scene}. Any pixels or annotations corresponding to the \texttt{background} class are simply ignored.

The quantitative results for foreground semantic segmentation are presented in \Cref{table:zeroshot_without_bkg}. To account for variations introduced by the inherent stochasticity of the optimization process, we report the mean and standard deviation over four independent runs. This approach helps avoid skewing the results with values that might simply be due to chance. The table’s first block of rows represents prior work, while the two subsequent blocks represent our experiments conducted on subsets of the LAION-400M~\cite{schuhmann2021laion} and COCO-Captions datasets~\cite{chen2015microsoft}, respectively. For each dataset, we compare three configurations: (1) the baseline SimZSS model \cite{simzss} trained from scratch on raw captions, (2) training from scratch using synthetic captions generated by the vision-language model (VLM), and (3) an additional fine-tuning step using the same alignment loss on all downstream training datasets. Notably, we observe significant gains in mean Intersection over Union (mIoU) across all datasets when synthetic captions from the VLM are used to train the text encoder. This improvement is even more pronounced on the subset of LAION, likely due to the higher noise level in image-caption pairs. Finally, the additional fine-tuning further improves the performance on average, regardless of the pretraining dataset.

The left side of \Cref{fig:laion_scaling} compares the performance of these same three approaches as we vary the subset size of LAION-400M~\cite{schuhmann2021laion}. The findings mirror those in \Cref{table:zeroshot_without_bkg}: training on synthetic captions and further fine-tuning on downstream datasets both lead to performance improvements.

\subsection{Whole Image Open-Vocabulary Zero-Shot Segmentation}
\label{subsec:openvoc-wholeimage}

Following the evaluation standard proposed by \cite{cha2023learning}, we include a zero-shot segmentation task that incorporates the \texttt{background} class. Rather than relying on a textual representation to identify \texttt{background} pixels, we apply a confidence-based strategy, classifying pixels as \texttt{background} when the model's confidence in any known class falls below a model- and dataset-specific threshold. It’s worth noting, however, that this evaluation has limitations. The benchmark favors models that are well-calibrated across the dataset’s classes, but this calibration can be misleading in an open-vocabulary context. For example, consider a dataset with only \texttt{dog}, \texttt{background}, and \texttt{cat} classes. A pixel actually representing \texttt{grass} may cause the model to incorrectly lean toward \texttt{cat} or \texttt{dog} if \texttt{grass} is missing from the label set. If \texttt{grass} were available as a label, the model might confidently label the pixel as \texttt{grass}. As such, this benchmark could unfairly penalize models capable of correctly handling unseen classes. While we include this evaluation for completeness, we recommend focusing on datasets where every pixel is fully labeled, such as Cityscapes~\cite{cordts2016cityscapes}. Other inference settings are aligned with those described in \Cref{subsec:openvoc}.

The quantitative results for whole image semantic segmentation are presented in \Cref{table:zeroshot_with_bkg}, which includes the same two bottom blocks of rows as \Cref{table:zeroshot_without_bkg}. Here, we observe similar trends: replacing raw captions with synthetic captions consistently boosts mIoU performance across all datasets. On the LAION-10M subset, performance further improves with additional fine-tuning, while on COCO-Captions, the mIoU remains unchanged. We speculate that this finetuning may be increasing the model’s confidence on classes often labeled as \texttt{background}, improving the segmentation accuracy for specific classes but potentially reducing performance for pixels labeled as \texttt{background}.

The right side of \Cref{fig:laion_scaling} shows further comparison of the 3 approaches, with varying sizes of the LAION-400 subset. The conclusions are analogous to those in \Cref{table:zeroshot_with_bkg}, both the training on synthetic captions as well as the finetuning on the downstream datasets improves the performance.

\input{tables_tex/ablation_vlm}
\input{tables_tex/training_stat}

\subsection{Ablations}
\label{subsec:ablations}
The only additional hyperparameter our method introduces is the prompt used in the VLM to generate the synthetic captions. An ablation over different prompts is shown in \Cref{table:caption_ablation}. The prompts were chosen in a way for the VLM to generate a concise image description. Regardless of the prompt used, the VLM consistently generated synthetic captions that improved the zero-shot open-vocabulary segmentation. Interestingly, since most prompts demonstrated comparable performance, a dummy prompt, \texttt{Write a story about aliens}, was introduced to verify that the choice of prompt indeed affects zero-shot segmentation performance. 

Additionally, different VLMs were tested, as shown in \Cref{table:vlm_ablation}, yielding results akin to those in \Cref{table:caption_ablation}. Regardless of the VLM, the use of synthetic captions led to improved performance. Further experimentation with more powerful VLMs may lead to additional performance improvements and is left as future work.

\subsection{Empirical Observations}
\label{subsec:empirical-observations}

\subsubsection{Training metrics}
In the quest to shed some light on \textit{``Why are synthetic captions much better than raw captions?''}, we look into the occurrences of class labels of the downstream datasets in the captions. \Cref{table:training_statistics} shows 1) the average number concept identified per caption and 2) the average number of concepts identified per batch. It can be observed that both metrics are significantly higher when using the synthetic captions and are most likely part of the reason why the the synthetic captions outperform the baseline with raw captions. In particular, it can be observed that synthetic captions contain on average $20\times$ the number of concepts compared to raw captions.

\subsubsection{Qualitative dense vision-language alignment}
\Cref{fig:heatmap} shows a qualitative visualization of the dense vision-language alignment. The heatmaps are obtained by computing the cosine similarity between a patch-representation and the representation of a query textual concept. It can be observed that the alignment is coherent, even on small objects showcasing the granularity of the vision-language alignment. This visualization is showed at $4\times$ the training resolution.

\input{figures_tex/heatmap}

%% file: tables_tex/ablation_prompt.tex
\begin{table*}[t]
\centering
\caption{
\textbf{Ablation over different prompts.} The ablation is performed on a 600k subset of LAION-400M for 6 epochs.
}
\resizebox{1\textwidth}{!}{
\begin{tabular}{lcc}
\toprule
\textbf{Configuration} & \textbf{Avg. foreground} & \textbf{Avg. whole image} \\
\midrule
\textcolor{gray}{Baseline} \\
Original image-caption pairs & $31.8 \pm 0.58$ & $33.8 \pm 0.36$ \\

\arrayrulecolor{black!15}\midrule[0.25pt]
\textcolor{gray}{Prompts} \\
\texttt{"Write a story about aliens."} & $5.8\pm1.35$ & $4.5\pm1.19$\\
\texttt{"What do you see?"} & $37.2\pm0.33$ & $38.5\pm0.61$ \\
\texttt{"What objects are present in the image?"} & $37.2\pm0.47$ & $39.3\pm0.63$ \\
\texttt{""} & $37.2\pm1.01$ & $38.6\pm1.25$\\
\texttt{"Describe the image. Only give descriptions you are 100\% certain of."} & $37.5 \pm 0.15$ & $39.0 \pm 0.43$ \\
\texttt{"Describe the image."} & $37.7 \pm 0.73$ & $39.0 \pm 0.97$ \\
\rowcolor{light_cyan} \texttt{"Very briefly describe the image."} & $39.1 \pm 0.56$ & $40.3 \pm 0.83$ \\

\end{tabular}
\label{table:caption_ablation}
}

\end{table*}

%% file: tables_tex/ablation_vlm.tex
\begin{table}[t]
\centering
\caption{
\textbf{Ablation over different VLMs.} For all VLMs, the prompt used is \texttt{"Very briefly describe the image."}. The alignment is performed on a ~600k subset of LAION-400M for 6 epochs. The synthetic captions perform better than the baseline irrespective of the VLM.
}
\resizebox{1\columnwidth}{!}{
\begin{tabular}{lcc}
\toprule
\textbf{Configuration} & \textbf{Avg. foreground} & \textbf{Avg. whole image} \\
\midrule
\textcolor{gray}{Baseline} \\
Original image-caption pairs & $31.8 \pm 0.58$ & $33.8 \pm 0.36$ \\
\arrayrulecolor{black!15}\midrule[0.25pt]
\textcolor{gray}{VLMs} \\
LLaVA-NeXT-Llama-3.1-8B & $37.8\pm0.12$ & $39.7 \pm 0.97$ \\
\rowcolor{light_cyan} LLaVA-NeXT-Mistral-7B & $39.1 \pm 0.56$ & $40.3 \pm 0.83$ \\
\end{tabular}
\label{table:vlm_ablation}
}

\end{table}

%% file: tables_tex/training_stat.tex
\begin{table}[t]
\centering
\caption{
\textbf{Alignment training statistics.} The alignment is performed on a ~600k subset of LAION-400M with a batchsize of 16384.
}
\resizebox{1\columnwidth}{!}{
\begin{tabular}{lcc}
\toprule
\textbf{Configuration} & \textbf{\#concept / caption} & \textbf{\#unique concept / batch} \\
\midrule
Baseline & $\sim$0.08 & $\sim$300 \\
\rowcolor{light_cyan} Baseline w/ synthetic captions & $\sim$2.3 & $\sim$430 \\
\end{tabular}
\label{table:training_statistics}
}

\end{table}

%% file: figures_tex/heatmap.tex
\begin{figure}[t]
  \centering
  \includegraphics[width=\linewidth]{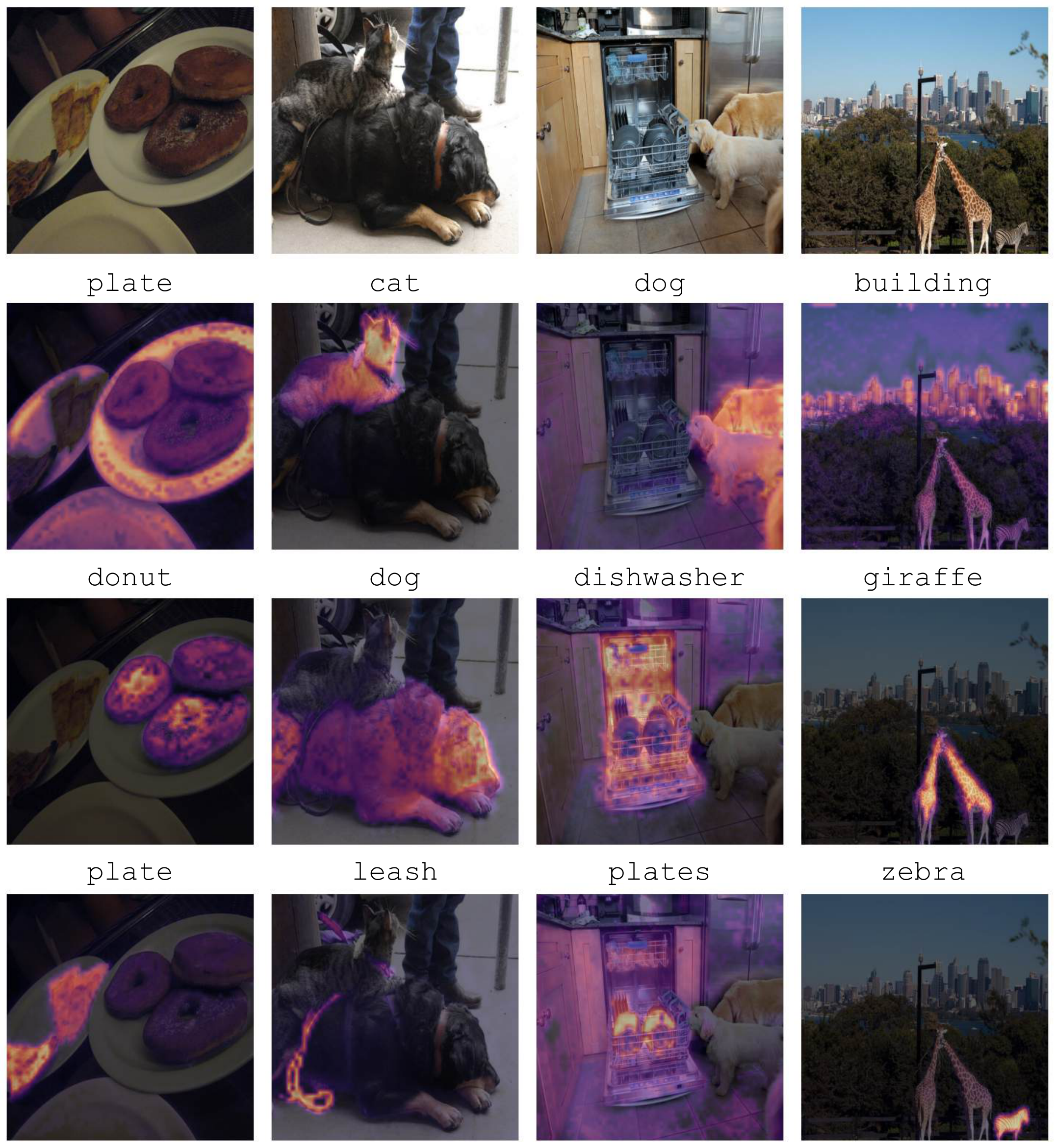}
  \caption{\textbf{Qualitative visualization of the dense vision-language alignment.} Each heatmap shows the pairwise similarities between pixel patch representations and a query textual concept.}
  \label{fig:heatmap}
\end{figure}

%% file: sec/conclusion.tex
\section{Conclusion}
In conclusion, we introduce a dense vision-language alignment method that exclusively utilizes synthetic captions generated by generative vision-language models (VLMs). This approach leverages the high-level understanding of generative VLMs to produce synthetic image descriptions which are used for explicit dense vision-language alignment. By combining these strategies, we successfully remove the reliance on traditional image-caption pairs, thereby establishing a more consistently defined alignment objective. The efficacy of our method is validated through standard zero-shot open-vocabulary benchmarks and qualitative assessments of the dense vision-language alignment. Additionally, our method offers scaling advantages, as improvements in VLMs will directly enhance our approach. Future work will focus on directly training the VLM in a similar manner to consolidate both 1) the image-level understanding and 2) the dense vision-language alignment into a single, unified model.

\vspace{0.2cm}
\noindent \textbf{Limitations} \\
While synthetic captions greatly improve the model’s ability to learn dense vision-language alignment, generating these captions involves an additional forward pass through the VLM, which can be computationally expensive. To address this, we precompute and store the synthetic captions on disk after their initial generation. This approach allows us to amortize the inference cost over multiple alignment training epochs, significantly reducing the computational overhead in practice.

%% file: main.bbl
\begin{thebibliography}{53}
\providecommand{\natexlab}[1]{#1}
\providecommand{\url}[1]{\texttt{#1}}
\expandafter\ifx\csname urlstyle\endcsname\relax
  \providecommand{\doi}[1]{doi: #1}\else
  \providecommand{\doi}{doi: \begingroup \urlstyle{rm}\Url}\fi

\bibitem[Alayrac et~al.(2022)Alayrac, Donahue, Luc, Miech, Barr, Hasson, Lenc, Mensch, Millican, Reynolds, Ring, Rutherford, Cabi, Han, Gong, Samangooei, Monteiro, Menick, Borgeaud, Brock, Nematzadeh, Sharifzadeh, Binkowski, Barreira, Vinyals, Zisserman, and Simonyan]{alayrac2022flamingovisuallanguagemodel}
Jean-Baptiste Alayrac, Jeff Donahue, Pauline Luc, Antoine Miech, Iain Barr, Yana Hasson, Karel Lenc, Arthur Mensch, Katie Millican, Malcolm Reynolds, Roman Ring, Eliza Rutherford, Serkan Cabi, Tengda Han, Zhitao Gong, Sina Samangooei, Marianne Monteiro, Jacob Menick, Sebastian Borgeaud, Andrew Brock, Aida Nematzadeh, Sahand Sharifzadeh, Mikolaj Binkowski, Ricardo Barreira, Oriol Vinyals, Andrew Zisserman, and Karen Simonyan.
\newblock Flamingo: a visual language model for few-shot learning, 2022.

\bibitem[Araslanov and Roth(2020)]{araslanov2020single}
Nikita Araslanov and Stefan Roth.
\newblock Single-stage semantic segmentation from image labels.
\newblock In \emph{Proceedings of the IEEE/CVF conference on computer vision and pattern recognition}, pages 4253--4262, 2020.

\bibitem[Caesar et~al.(2018)Caesar, Uijlings, and Ferrari]{caesar2018coco}
Holger Caesar, Jasper Uijlings, and Vittorio Ferrari.
\newblock Coco-stuff: Thing and stuff classes in context.
\newblock In \emph{Proceedings of the IEEE conference on computer vision and pattern recognition}, pages 1209--1218, 2018.

\bibitem[Caron et~al.(2021)Caron, Touvron, Misra, J{\'e}gou, Mairal, Bojanowski, and Joulin]{caron2021emerging}
Mathilde Caron, Hugo Touvron, Ishan Misra, Herv{\'e} J{\'e}gou, Julien Mairal, Piotr Bojanowski, and Armand Joulin.
\newblock Emerging properties in self-supervised vision transformers.
\newblock In \emph{Proceedings of the IEEE/CVF international conference on computer vision}, pages 9650--9660, 2021.

\bibitem[Cha et~al.(2023)Cha, Mun, and Roh]{cha2023learning}
Junbum Cha, Jonghwan Mun, and Byungseok Roh.
\newblock Learning to generate text-grounded mask for open-world semantic segmentation from only image-text pairs.
\newblock In \emph{Proceedings of the IEEE/CVF Conference on Computer Vision and Pattern Recognition}, pages 11165--11174, 2023.

\bibitem[Chen et~al.(2023)Chen, Zhu, Qian, Ghanem, Yan, Zhu, Xiao, Culatana, and Elhoseiny]{chen2023exploring}
Jun Chen, Deyao Zhu, Guocheng Qian, Bernard Ghanem, Zhicheng Yan, Chenchen Zhu, Fanyi Xiao, Sean~Chang Culatana, and Mohamed Elhoseiny.
\newblock Exploring open-vocabulary semantic segmentation from clip vision encoder distillation only.
\newblock In \emph{Proceedings of the IEEE/CVF International Conference on Computer Vision}, pages 699--710, 2023.

\bibitem[Chen et~al.(2020)Chen, Kornblith, Norouzi, and Hinton]{chen2020simple}
Ting Chen, Simon Kornblith, Mohammad Norouzi, and Geoffrey Hinton.
\newblock A simple framework for contrastive learning of visual representations.
\newblock In \emph{International conference on machine learning}, pages 1597--1607. PMLR, 2020.

\bibitem[Chen et~al.(2015)Chen, Fang, Lin, Vedantam, Gupta, Doll{\'a}r, and Zitnick]{chen2015microsoft}
Xinlei Chen, Hao Fang, Tsung-Yi Lin, Ramakrishna Vedantam, Saurabh Gupta, Piotr Doll{\'a}r, and C~Lawrence Zitnick.
\newblock Microsoft coco captions: Data collection and evaluation server.
\newblock \emph{arXiv preprint arXiv:1504.00325}, 2015.

\bibitem[Contributors(2020)]{mmseg2020}
MMSegmentation Contributors.
\newblock {MMSegmentation}: Openmmlab semantic segmentation toolbox and benchmark.
\newblock \url{https://github.com/open-mmlab/mmsegmentation}, 2020.

\bibitem[Cordts et~al.(2016)Cordts, Omran, Ramos, Rehfeld, Enzweiler, Benenson, Franke, Roth, and Schiele]{cordts2016cityscapes}
Marius Cordts, Mohamed Omran, Sebastian Ramos, Timo Rehfeld, Markus Enzweiler, Rodrigo Benenson, Uwe Franke, Stefan Roth, and Bernt Schiele.
\newblock The cityscapes dataset for semantic urban scene understanding.
\newblock In \emph{Proceedings of the IEEE conference on computer vision and pattern recognition}, pages 3213--3223, 2016.

\bibitem[Dong et~al.(2023)Dong, Bao, Zheng, Zhang, Chen, Yang, Zeng, Zhang, Yuan, Chen, Wen, and Yu]{Dong_2023_CVPR}
Xiaoyi Dong, Jianmin Bao, Yinglin Zheng, Ting Zhang, Dongdong Chen, Hao Yang, Ming Zeng, Weiming Zhang, Lu Yuan, Dong Chen, Fang Wen, and Nenghai Yu.
\newblock Maskclip: Masked self-distillation advances contrastive language-image pretraining.
\newblock In \emph{Proceedings of the IEEE/CVF Conference on Computer Vision and Pattern Recognition (CVPR)}, pages 10995--11005, 2023.

\bibitem[Dubey et~al.(2024)Dubey, Jauhri, Pandey, Kadian, Al-Dahle, Letman, Mathur, Schelten, Yang, Fan, Goyal, Hartshorn, Yang, Mitra, Sravankumar, Korenev, Hinsvark, Rao, Zhang, Rodriguez, Gregerson, Spataru, Roziere, Biron, Tang, Chern, Caucheteux, Nayak, Bi, Marra, McConnell, Keller, Touret, Wu, Wong, Ferrer, Nikolaidis, Allonsius, Song, Pintz, Livshits, Esiobu, Choudhary, Mahajan, Garcia-Olano, Perino, Hupkes, Lakomkin, AlBadawy, Lobanova, Dinan, Smith, Radenovic, Zhang, Synnaeve, Lee, Anderson, Nail, Mialon, Pang, Cucurell, Nguyen, Korevaar, Xu, Touvron, Zarov, Ibarra, Kloumann, Misra, Evtimov, Copet, Lee, Geffert, Vranes, Park, Mahadeokar, Shah, van~der Linde, Billock, Hong, Lee, Fu, Chi, Huang, Liu, Wang, Yu, Bitton, Spisak, Park, Rocca, Johnstun, Saxe, Jia, Alwala, Upasani, Plawiak, Li, Heafield, Stone, El-Arini, Iyer, Malik, Chiu, Bhalla, Rantala-Yeary, van~der Maaten, Chen, Tan, Jenkins, Martin, Madaan, Malo, Blecher, Landzaat, de~Oliveira, Muzzi, Pasupuleti, Singh, Paluri, Kardas, Oldham, Rita,
  Pavlova, Kambadur, Lewis, Si, Singh, Hassan, Goyal, Torabi, Bashlykov, Bogoychev, Chatterji, Duchenne, Çelebi, Alrassy, Zhang, Li, Vasic, Weng, Bhargava, Dubal, Krishnan, Koura, Xu, He, Dong, Srinivasan, Ganapathy, Calderer, Cabral, Stojnic, Raileanu, Girdhar, Patel, Sauvestre, Polidoro, Sumbaly, Taylor, Silva, Hou, Wang, Hosseini, Chennabasappa, Singh, Bell, Kim, Edunov, Nie, Narang, Raparthy, Shen, Wan, Bhosale, Zhang, Vandenhende, Batra, Whitman, Sootla, Collot, Gururangan, Borodinsky, Herman, Fowler, Sheasha, Georgiou, Scialom, Speckbacher, Mihaylov, Xiao, Karn, Goswami, Gupta, Ramanathan, Kerkez, Gonguet, Do, Vogeti, Petrovic, Chu, Xiong, Fu, Meers, Martinet, Wang, Tan, Xie, Jia, Wang, Goldschlag, Gaur, Babaei, Wen, Song, Zhang, Li, Mao, Coudert, Yan, Chen, Papakipos, Singh, Grattafiori, Jain, Kelsey, Shajnfeld, Gangidi, Victoria, Goldstand, Menon, Sharma, Boesenberg, Vaughan, Baevski, Feinstein, Kallet, Sangani, Yunus, Lupu, Alvarado, Caples, Gu, Ho, Poulton, Ryan, Ramchandani, Franco, Saraf,
  Chowdhury, Gabriel, Bharambe, Eisenman, Yazdan, James, Maurer, Leonhardi, Huang, Loyd, Paola, Paranjape, Liu, Wu, Ni, Hancock, Wasti, Spence, Stojkovic, Gamido, Montalvo, Parker, Burton, Mejia, Wang, Kim, Zhou, Hu, Chu, Cai, Tindal, Feichtenhofer, Civin, Beaty, Kreymer, Li, Wyatt, Adkins, Xu, Testuggine, David, Parikh, Liskovich, Foss, Wang, Le, Holland, Dowling, Jamil, Montgomery, Presani, Hahn, Wood, Brinkman, Arcaute, Dunbar, Smothers, Sun, Kreuk, Tian, Ozgenel, Caggioni, Guzmán, Kanayet, Seide, Florez, Schwarz, Badeer, Swee, Halpern, Thattai, Herman, Sizov, Guangyi, Zhang, Lakshminarayanan, Shojanazeri, Zou, Wang, Zha, Habeeb, Rudolph, Suk, Aspegren, Goldman, Damlaj, Molybog, Tufanov, Veliche, Gat, Weissman, Geboski, Kohli, Asher, Gaya, Marcus, Tang, Chan, Zhen, Reizenstein, Teboul, Zhong, Jin, Yang, Cummings, Carvill, Shepard, McPhie, Torres, Ginsburg, Wang, Wu, U, Saxena, Prasad, Khandelwal, Zand, Matosich, Veeraraghavan, Michelena, Li, Huang, Chawla, Lakhotia, Huang, Chen, Garg, A, Silva, Bell,
  Zhang, Guo, Yu, Moshkovich, Wehrstedt, Khabsa, Avalani, Bhatt, Tsimpoukelli, Mankus, Hasson, Lennie, Reso, Groshev, Naumov, Lathi, Keneally, Seltzer, Valko, Restrepo, Patel, Vyatskov, Samvelyan, Clark, Macey, Wang, Hermoso, Metanat, Rastegari, Bansal, Santhanam, Parks, White, Bawa, Singhal, Egebo, Usunier, Laptev, Dong, Zhang, Cheng, Chernoguz, Hart, Salpekar, Kalinli, Kent, Parekh, Saab, Balaji, Rittner, Bontrager, Roux, Dollar, Zvyagina, Ratanchandani, Yuvraj, Liang, Alao, Rodriguez, Ayub, Murthy, Nayani, Mitra, Li, Hogan, Battey, Wang, Maheswari, Howes, Rinott, Bondu, Datta, Chugh, Hunt, Dhillon, Sidorov, Pan, Verma, Yamamoto, Ramaswamy, Lindsay, Lindsay, Feng, Lin, Zha, Shankar, Zhang, Zhang, Wang, Agarwal, Sajuyigbe, Chintala, Max, Chen, Kehoe, Satterfield, Govindaprasad, Gupta, Cho, Virk, Subramanian, Choudhury, Goldman, Remez, Glaser, Best, Kohler, Robinson, Li, Zhang, Matthews, Chou, Shaked, Vontimitta, Ajayi, Montanez, Mohan, Kumar, Mangla, Albiero, Ionescu, Poenaru, Mihailescu, Ivanov, Li, Wang,
  Jiang, Bouaziz, Constable, Tang, Wang, Wu, Wang, Xia, Wu, Gao, Chen, Hu, Jia, Qi, Li, Zhang, Zhang, Adi, Nam, Yu, Wang, Hao, Qian, He, Rait, DeVito, Rosnbrick, Wen, Yang, and Zhao]{dubey2024llama3herdmodels}
Abhimanyu Dubey, Abhinav Jauhri, Abhinav Pandey, Abhishek Kadian, Ahmad Al-Dahle, Aiesha Letman, Akhil Mathur, Alan Schelten, Amy Yang, Angela Fan, Anirudh Goyal, Anthony Hartshorn, Aobo Yang, Archi Mitra, Archie Sravankumar, Artem Korenev, Arthur Hinsvark, Arun Rao, Aston Zhang, Aurelien Rodriguez, Austen Gregerson, Ava Spataru, Baptiste Roziere, Bethany Biron, Binh Tang, Bobbie Chern, Charlotte Caucheteux, Chaya Nayak, Chloe Bi, Chris Marra, Chris McConnell, Christian Keller, Christophe Touret, Chunyang Wu, Corinne Wong, Cristian~Canton Ferrer, Cyrus Nikolaidis, Damien Allonsius, Daniel Song, Danielle Pintz, Danny Livshits, David Esiobu, Dhruv Choudhary, Dhruv Mahajan, Diego Garcia-Olano, Diego Perino, Dieuwke Hupkes, Egor Lakomkin, Ehab AlBadawy, Elina Lobanova, Emily Dinan, Eric~Michael Smith, Filip Radenovic, Frank Zhang, Gabriel Synnaeve, Gabrielle Lee, Georgia~Lewis Anderson, Graeme Nail, Gregoire Mialon, Guan Pang, Guillem Cucurell, Hailey Nguyen, Hannah Korevaar, Hu Xu, Hugo Touvron, Iliyan Zarov,
  Imanol~Arrieta Ibarra, Isabel Kloumann, Ishan Misra, Ivan Evtimov, Jade Copet, Jaewon Lee, Jan Geffert, Jana Vranes, Jason Park, Jay Mahadeokar, Jeet Shah, Jelmer van~der Linde, Jennifer Billock, Jenny Hong, Jenya Lee, Jeremy Fu, Jianfeng Chi, Jianyu Huang, Jiawen Liu, Jie Wang, Jiecao Yu, Joanna Bitton, Joe Spisak, Jongsoo Park, Joseph Rocca, Joshua Johnstun, Joshua Saxe, Junteng Jia, Kalyan~Vasuden Alwala, Kartikeya Upasani, Kate Plawiak, Ke Li, Kenneth Heafield, Kevin Stone, Khalid El-Arini, Krithika Iyer, Kshitiz Malik, Kuenley Chiu, Kunal Bhalla, Lauren Rantala-Yeary, Laurens van~der Maaten, Lawrence Chen, Liang Tan, Liz Jenkins, Louis Martin, Lovish Madaan, Lubo Malo, Lukas Blecher, Lukas Landzaat, Luke de Oliveira, Madeline Muzzi, Mahesh Pasupuleti, Mannat Singh, Manohar Paluri, Marcin Kardas, Mathew Oldham, Mathieu Rita, Maya Pavlova, Melanie Kambadur, Mike Lewis, Min Si, Mitesh~Kumar Singh, Mona Hassan, Naman Goyal, Narjes Torabi, Nikolay Bashlykov, Nikolay Bogoychev, Niladri Chatterji, Olivier
  Duchenne, Onur Çelebi, Patrick Alrassy, Pengchuan Zhang, Pengwei Li, Petar Vasic, Peter Weng, Prajjwal Bhargava, Pratik Dubal, Praveen Krishnan, Punit~Singh Koura, Puxin Xu, Qing He, Qingxiao Dong, Ragavan Srinivasan, Raj Ganapathy, Ramon Calderer, Ricardo~Silveira Cabral, Robert Stojnic, Roberta Raileanu, Rohit Girdhar, Rohit Patel, Romain Sauvestre, Ronnie Polidoro, Roshan Sumbaly, Ross Taylor, Ruan Silva, Rui Hou, Rui Wang, Saghar Hosseini, Sahana Chennabasappa, Sanjay Singh, Sean Bell, Seohyun~Sonia Kim, Sergey Edunov, Shaoliang Nie, Sharan Narang, Sharath Raparthy, Sheng Shen, Shengye Wan, Shruti Bhosale, Shun Zhang, Simon Vandenhende, Soumya Batra, Spencer Whitman, Sten Sootla, Stephane Collot, Suchin Gururangan, Sydney Borodinsky, Tamar Herman, Tara Fowler, Tarek Sheasha, Thomas Georgiou, Thomas Scialom, Tobias Speckbacher, Todor Mihaylov, Tong Xiao, Ujjwal Karn, Vedanuj Goswami, Vibhor Gupta, Vignesh Ramanathan, Viktor Kerkez, Vincent Gonguet, Virginie Do, Vish Vogeti, Vladan Petrovic, Weiwei Chu,
  Wenhan Xiong, Wenyin Fu, Whitney Meers, Xavier Martinet, Xiaodong Wang, Xiaoqing~Ellen Tan, Xinfeng Xie, Xuchao Jia, Xuewei Wang, Yaelle Goldschlag, Yashesh Gaur, Yasmine Babaei, Yi Wen, Yiwen Song, Yuchen Zhang, Yue Li, Yuning Mao, Zacharie~Delpierre Coudert, Zheng Yan, Zhengxing Chen, Zoe Papakipos, Aaditya Singh, Aaron Grattafiori, Abha Jain, Adam Kelsey, Adam Shajnfeld, Adithya Gangidi, Adolfo Victoria, Ahuva Goldstand, Ajay Menon, Ajay Sharma, Alex Boesenberg, Alex Vaughan, Alexei Baevski, Allie Feinstein, Amanda Kallet, Amit Sangani, Anam Yunus, Andrei Lupu, Andres Alvarado, Andrew Caples, Andrew Gu, Andrew Ho, Andrew Poulton, Andrew Ryan, Ankit Ramchandani, Annie Franco, Aparajita Saraf, Arkabandhu Chowdhury, Ashley Gabriel, Ashwin Bharambe, Assaf Eisenman, Azadeh Yazdan, Beau James, Ben Maurer, Benjamin Leonhardi, Bernie Huang, Beth Loyd, Beto~De Paola, Bhargavi Paranjape, Bing Liu, Bo Wu, Boyu Ni, Braden Hancock, Bram Wasti, Brandon Spence, Brani Stojkovic, Brian Gamido, Britt Montalvo, Carl
  Parker, Carly Burton, Catalina Mejia, Changhan Wang, Changkyu Kim, Chao Zhou, Chester Hu, Ching-Hsiang Chu, Chris Cai, Chris Tindal, Christoph Feichtenhofer, Damon Civin, Dana Beaty, Daniel Kreymer, Daniel Li, Danny Wyatt, David Adkins, David Xu, Davide Testuggine, Delia David, Devi Parikh, Diana Liskovich, Didem Foss, Dingkang Wang, Duc Le, Dustin Holland, Edward Dowling, Eissa Jamil, Elaine Montgomery, Eleonora Presani, Emily Hahn, Emily Wood, Erik Brinkman, Esteban Arcaute, Evan Dunbar, Evan Smothers, Fei Sun, Felix Kreuk, Feng Tian, Firat Ozgenel, Francesco Caggioni, Francisco Guzmán, Frank Kanayet, Frank Seide, Gabriela~Medina Florez, Gabriella Schwarz, Gada Badeer, Georgia Swee, Gil Halpern, Govind Thattai, Grant Herman, Grigory Sizov, Guangyi, Zhang, Guna Lakshminarayanan, Hamid Shojanazeri, Han Zou, Hannah Wang, Hanwen Zha, Haroun Habeeb, Harrison Rudolph, Helen Suk, Henry Aspegren, Hunter Goldman, Ibrahim Damlaj, Igor Molybog, Igor Tufanov, Irina-Elena Veliche, Itai Gat, Jake Weissman, James
  Geboski, James Kohli, Japhet Asher, Jean-Baptiste Gaya, Jeff Marcus, Jeff Tang, Jennifer Chan, Jenny Zhen, Jeremy Reizenstein, Jeremy Teboul, Jessica Zhong, Jian Jin, Jingyi Yang, Joe Cummings, Jon Carvill, Jon Shepard, Jonathan McPhie, Jonathan Torres, Josh Ginsburg, Junjie Wang, Kai Wu, Kam~Hou U, Karan Saxena, Karthik Prasad, Kartikay Khandelwal, Katayoun Zand, Kathy Matosich, Kaushik Veeraraghavan, Kelly Michelena, Keqian Li, Kun Huang, Kunal Chawla, Kushal Lakhotia, Kyle Huang, Lailin Chen, Lakshya Garg, Lavender A, Leandro Silva, Lee Bell, Lei Zhang, Liangpeng Guo, Licheng Yu, Liron Moshkovich, Luca Wehrstedt, Madian Khabsa, Manav Avalani, Manish Bhatt, Maria Tsimpoukelli, Martynas Mankus, Matan Hasson, Matthew Lennie, Matthias Reso, Maxim Groshev, Maxim Naumov, Maya Lathi, Meghan Keneally, Michael~L. Seltzer, Michal Valko, Michelle Restrepo, Mihir Patel, Mik Vyatskov, Mikayel Samvelyan, Mike Clark, Mike Macey, Mike Wang, Miquel~Jubert Hermoso, Mo Metanat, Mohammad Rastegari, Munish Bansal, Nandhini
  Santhanam, Natascha Parks, Natasha White, Navyata Bawa, Nayan Singhal, Nick Egebo, Nicolas Usunier, Nikolay~Pavlovich Laptev, Ning Dong, Ning Zhang, Norman Cheng, Oleg Chernoguz, Olivia Hart, Omkar Salpekar, Ozlem Kalinli, Parkin Kent, Parth Parekh, Paul Saab, Pavan Balaji, Pedro Rittner, Philip Bontrager, Pierre Roux, Piotr Dollar, Polina Zvyagina, Prashant Ratanchandani, Pritish Yuvraj, Qian Liang, Rachad Alao, Rachel Rodriguez, Rafi Ayub, Raghotham Murthy, Raghu Nayani, Rahul Mitra, Raymond Li, Rebekkah Hogan, Robin Battey, Rocky Wang, Rohan Maheswari, Russ Howes, Ruty Rinott, Sai~Jayesh Bondu, Samyak Datta, Sara Chugh, Sara Hunt, Sargun Dhillon, Sasha Sidorov, Satadru Pan, Saurabh Verma, Seiji Yamamoto, Sharadh Ramaswamy, Shaun Lindsay, Shaun Lindsay, Sheng Feng, Shenghao Lin, Shengxin~Cindy Zha, Shiva Shankar, Shuqiang Zhang, Shuqiang Zhang, Sinong Wang, Sneha Agarwal, Soji Sajuyigbe, Soumith Chintala, Stephanie Max, Stephen Chen, Steve Kehoe, Steve Satterfield, Sudarshan Govindaprasad, Sumit Gupta,
  Sungmin Cho, Sunny Virk, Suraj Subramanian, Sy Choudhury, Sydney Goldman, Tal Remez, Tamar Glaser, Tamara Best, Thilo Kohler, Thomas Robinson, Tianhe Li, Tianjun Zhang, Tim Matthews, Timothy Chou, Tzook Shaked, Varun Vontimitta, Victoria Ajayi, Victoria Montanez, Vijai Mohan, Vinay~Satish Kumar, Vishal Mangla, Vítor Albiero, Vlad Ionescu, Vlad Poenaru, Vlad~Tiberiu Mihailescu, Vladimir Ivanov, Wei Li, Wenchen Wang, Wenwen Jiang, Wes Bouaziz, Will Constable, Xiaocheng Tang, Xiaofang Wang, Xiaojian Wu, Xiaolan Wang, Xide Xia, Xilun Wu, Xinbo Gao, Yanjun Chen, Ye Hu, Ye Jia, Ye Qi, Yenda Li, Yilin Zhang, Ying Zhang, Yossi Adi, Youngjin Nam, Yu, Wang, Yuchen Hao, Yundi Qian, Yuzi He, Zach Rait, Zachary DeVito, Zef Rosnbrick, Zhaoduo Wen, Zhenyu Yang, and Zhiwei Zhao.
\newblock The llama 3 herd of models, 2024.

\bibitem[Everingham et~al.(2012)Everingham, Van~Gool, Williams, Winn, and Zisserman]{pascal-voc-2012}
M. Everingham, L. Van~Gool, C.~K.~I. Williams, J. Winn, and A. Zisserman.
\newblock The {PASCAL} {V}isual {O}bject {C}lasses {C}hallenge 2012 {(VOC2012)} {R}esults.
\newblock http://www.pascal-network.org/challenges/VOC/voc2012/workshop/index.html, 2012.

\bibitem[Gutmann and Hyvärinen(2010)]{nce_loss}
Michael Gutmann and Aapo Hyvärinen.
\newblock Noise-contrastive estimation: A new estimation principle for unnormalized statistical models.
\newblock In \emph{Proceedings of the Thirteenth International Conference on Artificial Intelligence and Statistics}, pages 297--304, Chia Laguna Resort, Sardinia, Italy, 2010. PMLR.

\bibitem[He et~al.(2017)He, Gkioxari, Doll{\'{a}}r, and Girshick]{DBLP:journals/corr/HeGDG17}
Kaiming He, Georgia Gkioxari, Piotr Doll{\'{a}}r, and Ross~B. Girshick.
\newblock Mask {R-CNN}.
\newblock \emph{CoRR}, abs/1703.06870, 2017.

\bibitem[He et~al.(2020)He, Fan, Wu, Xie, and Girshick]{he2020momentum}
Kaiming He, Haoqi Fan, Yuxin Wu, Saining Xie, and Ross Girshick.
\newblock Momentum contrast for unsupervised visual representation learning.
\newblock In \emph{Proceedings of the IEEE/CVF Conference on Computer Vision and Pattern Recognition}, pages 9729--9738, 2020.

\bibitem[Hjelm et~al.(2018)Hjelm, Fedorov, Lavoie-Marchildon, Grewal, Bachman, Trischler, and Bengio]{hjelm2018learning}
R~Devon Hjelm, Alex Fedorov, Samuel Lavoie-Marchildon, Karan Grewal, Phil Bachman, Adam Trischler, and Yoshua Bengio.
\newblock Learning deep representations by mutual information estimation and maximization.
\newblock \emph{arXiv preprint arXiv:1808.06670}, 2018.

\bibitem[Jia et~al.(2021)Jia, Yang, Xia, Chen, Parekh, Pham, Le, Sung, Li, and Duerig]{jia2021scaling}
Chao Jia, Yinfei Yang, Ye Xia, Yi-Ting Chen, Zarana Parekh, Hieu Pham, Quoc Le, Yun-Hsuan Sung, Zhen Li, and Tom Duerig.
\newblock Scaling up visual and vision-language representation learning with noisy text supervision.
\newblock In \emph{International conference on machine learning}, pages 4904--4916. PMLR, 2021.

\bibitem[Jiang et~al.(2023)Jiang, Sablayrolles, Mensch, Bamford, Chaplot, de~las Casas, Bressand, Lengyel, Lample, Saulnier, Lavaud, Lachaux, Stock, Scao, Lavril, Wang, Lacroix, and Sayed]{jiang2023mistral7b}
Albert~Q. Jiang, Alexandre Sablayrolles, Arthur Mensch, Chris Bamford, Devendra~Singh Chaplot, Diego de~las Casas, Florian Bressand, Gianna Lengyel, Guillaume Lample, Lucile Saulnier, Lélio~Renard Lavaud, Marie-Anne Lachaux, Pierre Stock, Teven~Le Scao, Thibaut Lavril, Thomas Wang, Timothée Lacroix, and William~El Sayed.
\newblock Mistral 7b, 2023.

\bibitem[Karazija et~al.(2023)Karazija, Laina, Vedaldi, and Rupprecht]{karazija2023diffusion}
Laurynas Karazija, Iro Laina, Andrea Vedaldi, and Christian Rupprecht.
\newblock Diffusion models for zero-shot open-vocabulary segmentation.
\newblock \emph{arXiv preprint arXiv:2306.09316}, 2023.

\bibitem[Kingma and Ba(2014)]{kingma2014adam}
Diederik~P Kingma and Jimmy Ba.
\newblock Adam: A method for stochastic optimization.
\newblock \emph{arXiv preprint arXiv:1412.6980}, 2014.

\bibitem[Kirillov et~al.(2023)Kirillov, Mintun, Ravi, Mao, Rolland, Gustafson, Xiao, Whitehead, Berg, Lo, Dollár, and Girshick]{segmentanything}
Alexander Kirillov, Eric Mintun, Nikhila Ravi, Hanzi Mao, Chloe Rolland, Laura Gustafson, Tete Xiao, Spencer Whitehead, Alexander~C. Berg, Wan-Yen Lo, Piotr Dollár, and Ross Girshick.
\newblock Segment anything, 2023.

\bibitem[Kr{\"{a}}henb{\"{u}}hl and Koltun(2012)]{densecrf}
Philipp Kr{\"{a}}henb{\"{u}}hl and Vladlen Koltun.
\newblock Efficient inference in fully connected crfs with gaussian edge potentials.
\newblock \emph{CoRR}, abs/1210.5644, 2012.

\bibitem[Lavoie et~al.(2024)Lavoie, Kirichenko, Ibrahim, Assran, Wilson, Courville, and Ballas]{lavoie2024modelingcaptiondiversitycontrastive}
Samuel Lavoie, Polina Kirichenko, Mark Ibrahim, Mahmoud Assran, Andrew~Gordon Wilson, Aaron Courville, and Nicolas Ballas.
\newblock Modeling caption diversity in contrastive vision-language pretraining, 2024.

\bibitem[Li et~al.(2022)Li, Li, Xiong, and Hoi]{DBLP:journals/corr/abs-2201-12086}
Junnan Li, Dongxu Li, Caiming Xiong, and Steven C.~H. Hoi.
\newblock {BLIP:} bootstrapping language-image pre-training for unified vision-language understanding and generation.
\newblock \emph{CoRR}, abs/2201.12086, 2022.

\bibitem[Li et~al.(2023)Li, Li, Savarese, and Hoi]{li2023blip2bootstrappinglanguageimagepretraining}
Junnan Li, Dongxu Li, Silvio Savarese, and Steven Hoi.
\newblock Blip-2: Bootstrapping language-image pre-training with frozen image encoders and large language models, 2023.

\bibitem[Liu et~al.(2023)Liu, Li, Wu, and Lee]{liu2023llava}
Haotian Liu, Chunyuan Li, Qingyang Wu, and Yong~Jae Lee.
\newblock Visual instruction tuning, 2023.

\bibitem[Liu et~al.(2024)Liu, Li, Li, Li, Zhang, Shen, and Lee]{liu2024llavanext}
Haotian Liu, Chunyuan Li, Yuheng Li, Bo Li, Yuanhan Zhang, Sheng Shen, and Yong~Jae Lee.
\newblock Llava-next: Improved reasoning, ocr, and world knowledge, 2024.

\bibitem[Luo et~al.(2023)Luo, Bao, Wu, He, and Li]{luo2023segclip}
Huaishao Luo, Junwei Bao, Youzheng Wu, Xiaodong He, and Tianrui Li.
\newblock Segclip: Patch aggregation with learnable centers for open-vocabulary semantic segmentation.
\newblock In \emph{International Conference on Machine Learning}, pages 23033--23044. PMLR, 2023.

\bibitem[Mottaghi et~al.(2014)Mottaghi, Chen, Liu, Cho, Lee, Fidler, Urtasun, and Yuille]{mottaghi2014role}
Roozbeh Mottaghi, Xianjie Chen, Xiaobai Liu, Nam-Gyu Cho, Seong-Whan Lee, Sanja Fidler, Raquel Urtasun, and Alan Yuille.
\newblock The role of context for object detection and semantic segmentation in the wild.
\newblock In \emph{Proceedings of the IEEE conference on computer vision and pattern recognition}, pages 891--898, 2014.

\bibitem[Mukhoti et~al.(2023)Mukhoti, Lin, Poursaeed, Wang, Shah, Torr, and Lim]{mukhoti2023open}
Jishnu Mukhoti, Tsung-Yu Lin, Omid Poursaeed, Rui Wang, Ashish Shah, Philip~HS Torr, and Ser-Nam Lim.
\newblock Open vocabulary semantic segmentation with patch aligned contrastive learning.
\newblock In \emph{Proceedings of the IEEE/CVF Conference on Computer Vision and Pattern Recognition}, pages 19413--19423, 2023.

\bibitem[Oquab et~al.(2023)Oquab, Darcet, Moutakanni, Vo, Szafraniec, Khalidov, Fernandez, Haziza, Massa, El-Nouby, Howes, Huang, Xu, Sharma, Li, Galuba, Rabbat, Assran, Ballas, Synnaeve, Misra, Jegou, Mairal, Labatut, Joulin, and Bojanowski]{oquab2023dinov2}
Maxime Oquab, Timothée Darcet, Theo Moutakanni, Huy~V. Vo, Marc Szafraniec, Vasil Khalidov, Pierre Fernandez, Daniel Haziza, Francisco Massa, Alaaeldin El-Nouby, Russell Howes, Po-Yao Huang, Hu Xu, Vasu Sharma, Shang-Wen Li, Wojciech Galuba, Mike Rabbat, Mido Assran, Nicolas Ballas, Gabriel Synnaeve, Ishan Misra, Herve Jegou, Julien Mairal, Patrick Labatut, Armand Joulin, and Piotr Bojanowski.
\newblock Dinov2: Learning robust visual features without supervision, 2023.

\bibitem[Radford et~al.(2021)Radford, Kim, Hallacy, Ramesh, Goh, Agarwal, Sastry, Askell, Mishkin, Clark, Krueger, and Sutskever]{radford2021learning}
Alec Radford, Jong~Wook Kim, Chris Hallacy, Aditya Ramesh, Gabriel Goh, Sandhini Agarwal, Girish Sastry, Amanda Askell, Pamela Mishkin, Jack Clark, Gretchen Krueger, and Ilya Sutskever.
\newblock Learning transferable visual models from natural language supervision, 2021.

\bibitem[Ranasinghe et~al.(2023)Ranasinghe, McKinzie, Ravi, Yang, Toshev, and Shlens]{ranasinghe2023perceptual}
Kanchana Ranasinghe, Brandon McKinzie, Sachin Ravi, Yinfei Yang, Alexander Toshev, and Jonathon Shlens.
\newblock Perceptual grouping in contrastive vision-language models.
\newblock In \emph{Proceedings of the IEEE/CVF International Conference on Computer Vision}, pages 5571--5584, 2023.

\bibitem[Rewatbowornwong et~al.(2023)Rewatbowornwong, Chatthee, Chuangsuwanich, and Suwajanakorn]{rewatbowornwong2023zero}
Pitchaporn Rewatbowornwong, Nattanat Chatthee, Ekapol Chuangsuwanich, and Supasorn Suwajanakorn.
\newblock Zero-guidance segmentation using zero segment labels.
\newblock In \emph{Proceedings of the IEEE/CVF International Conference on Computer Vision}, pages 1162--1172, 2023.

\bibitem[Ronneberger et~al.(2015)Ronneberger, Fischer, and Brox]{DBLP:journals/corr/RonnebergerFB15}
Olaf Ronneberger, Philipp Fischer, and Thomas Brox.
\newblock U-net: Convolutional networks for biomedical image segmentation.
\newblock \emph{CoRR}, abs/1505.04597, 2015.

\bibitem[Schuhmann et~al.(2021)Schuhmann, Vencu, Beaumont, Kaczmarczyk, Mullis, Katta, Coombes, Jitsev, and Komatsuzaki]{schuhmann2021laion}
Christoph Schuhmann, Richard Vencu, Romain Beaumont, Robert Kaczmarczyk, Clayton Mullis, Aarush Katta, Theo Coombes, Jenia Jitsev, and Aran Komatsuzaki.
\newblock Laion-400m: Open dataset of clip-filtered 400 million image-text pairs.
\newblock \emph{arXiv preprint arXiv:2111.02114}, 2021.

\bibitem[Shin et~al.(2022)Shin, Xie, and Albanie]{shin2022reco}
Gyungin Shin, Weidi Xie, and Samuel Albanie.
\newblock Reco: Retrieve and co-segment for zero-shot transfer.
\newblock \emph{Advances in Neural Information Processing Systems}, 35:\penalty0 33754--33767, 2022.

\bibitem[Stegm{\"u}ller et~al.(2025)Stegm{\"u}ller, Lebailly, Duki{\'c}, Bozorgtabar, Tuytelaars, and Thiran]{simzss}
Thomas Stegm{\"u}ller, Tim Lebailly, Nikola Duki{\'c}, Behzad Bozorgtabar, Tinne Tuytelaars, and Jean-Philippe Thiran.
\newblock A simple framework for open-vocabulary zero-shot segmentation.
\newblock In \emph{The Thirteenth International Conference on Learning Representations}, 2025.

\bibitem[Team(2024)]{chameleon}
Chameleon Team.
\newblock Chameleon: Mixed-modal early-fusion foundation models, 2024.

\bibitem[Touvron et~al.(2023{\natexlab{a}})Touvron, Lavril, Izacard, Martinet, Lachaux, Lacroix, Rozière, Goyal, Hambro, Azhar, Rodriguez, Joulin, Grave, and Lample]{touvron2023llamaopenefficientfoundation}
Hugo Touvron, Thibaut Lavril, Gautier Izacard, Xavier Martinet, Marie-Anne Lachaux, Timothée Lacroix, Baptiste Rozière, Naman Goyal, Eric Hambro, Faisal Azhar, Aurelien Rodriguez, Armand Joulin, Edouard Grave, and Guillaume Lample.
\newblock Llama: Open and efficient foundation language models, 2023{\natexlab{a}}.

\bibitem[Touvron et~al.(2023{\natexlab{b}})Touvron, Martin, Stone, Albert, Almahairi, Babaei, Bashlykov, Batra, Bhargava, Bhosale, Bikel, Blecher, Ferrer, Chen, Cucurull, Esiobu, Fernandes, Fu, Fu, Fuller, Gao, Goswami, Goyal, Hartshorn, Hosseini, Hou, Inan, Kardas, Kerkez, Khabsa, Kloumann, Korenev, Koura, Lachaux, Lavril, Lee, Liskovich, Lu, Mao, Martinet, Mihaylov, Mishra, Molybog, Nie, Poulton, Reizenstein, Rungta, Saladi, Schelten, Silva, Smith, Subramanian, Tan, Tang, Taylor, Williams, Kuan, Xu, Yan, Zarov, Zhang, Fan, Kambadur, Narang, Rodriguez, Stojnic, Edunov, and Scialom]{touvron2023llama2openfoundation}
Hugo Touvron, Louis Martin, Kevin Stone, Peter Albert, Amjad Almahairi, Yasmine Babaei, Nikolay Bashlykov, Soumya Batra, Prajjwal Bhargava, Shruti Bhosale, Dan Bikel, Lukas Blecher, Cristian~Canton Ferrer, Moya Chen, Guillem Cucurull, David Esiobu, Jude Fernandes, Jeremy Fu, Wenyin Fu, Brian Fuller, Cynthia Gao, Vedanuj Goswami, Naman Goyal, Anthony Hartshorn, Saghar Hosseini, Rui Hou, Hakan Inan, Marcin Kardas, Viktor Kerkez, Madian Khabsa, Isabel Kloumann, Artem Korenev, Punit~Singh Koura, Marie-Anne Lachaux, Thibaut Lavril, Jenya Lee, Diana Liskovich, Yinghai Lu, Yuning Mao, Xavier Martinet, Todor Mihaylov, Pushkar Mishra, Igor Molybog, Yixin Nie, Andrew Poulton, Jeremy Reizenstein, Rashi Rungta, Kalyan Saladi, Alan Schelten, Ruan Silva, Eric~Michael Smith, Ranjan Subramanian, Xiaoqing~Ellen Tan, Binh Tang, Ross Taylor, Adina Williams, Jian~Xiang Kuan, Puxin Xu, Zheng Yan, Iliyan Zarov, Yuchen Zhang, Angela Fan, Melanie Kambadur, Sharan Narang, Aurelien Rodriguez, Robert Stojnic, Sergey Edunov, and Thomas
  Scialom.
\newblock Llama 2: Open foundation and fine-tuned chat models, 2023{\natexlab{b}}.

\bibitem[Tsimpoukelli et~al.(2021)Tsimpoukelli, Menick, Cabi, Eslami, Vinyals, and Hill]{tsimpoukelli2021multimodal}
Maria Tsimpoukelli, Jacob Menick, Serkan Cabi, S.~M.~Ali Eslami, Oriol Vinyals, and Felix Hill.
\newblock Multimodal few-shot learning with frozen language models.
\newblock In \emph{Advances in Neural Information Processing Systems}, 2021.

\bibitem[van~den Oord et~al.(2018)van~den Oord, Li, and Vinyals]{infonce}
A{\"{a}}ron van~den Oord, Yazhe Li, and Oriol Vinyals.
\newblock Representation learning with contrastive predictive coding.
\newblock \emph{CoRR}, abs/1807.03748, 2018.

\bibitem[Wysocza{\'n}ska et~al.(2024)Wysocza{\'n}ska, Ramamonjisoa, Trzci{\'n}ski, and Sim{\'e}oni]{wysoczanska2024clip}
Monika Wysocza{\'n}ska, Micha{\"e}l Ramamonjisoa, Tomasz Trzci{\'n}ski, and Oriane Sim{\'e}oni.
\newblock Clip-diy: Clip dense inference yields open-vocabulary semantic segmentation for-free.
\newblock In \emph{Proceedings of the IEEE/CVF Winter Conference on Applications of Computer Vision}, pages 1403--1413, 2024.

\bibitem[Wysoczańska et~al.(2024)Wysoczańska, Siméoni, Ramamonjisoa, Bursuc, Trzciński, and Pérez]{wysoczańska2024clipdinoiser}
Monika Wysoczańska, Oriane Siméoni, Michaël Ramamonjisoa, Andrei Bursuc, Tomasz Trzciński, and Patrick Pérez.
\newblock Clip-dinoiser: Teaching clip a few dino tricks for open-vocabulary semantic segmentation, 2024.

\bibitem[Xu et~al.(2022)Xu, De~Mello, Liu, Byeon, Breuel, Kautz, and Wang]{xu2022groupvit}
Jiarui Xu, Shalini De~Mello, Sifei Liu, Wonmin Byeon, Thomas Breuel, Jan Kautz, and Xiaolong Wang.
\newblock Groupvit: Semantic segmentation emerges from text supervision.
\newblock \emph{arXiv preprint arXiv:2202.11094}, 2022.

\bibitem[Xu et~al.(2023)Xu, Hou, Zhang, Feng, Wang, Qiao, and Xie]{xu2023learning}
Jilan Xu, Junlin Hou, Yuejie Zhang, Rui Feng, Yi Wang, Yu Qiao, and Weidi Xie.
\newblock Learning open-vocabulary semantic segmentation models from natural language supervision.
\newblock In \emph{Proceedings of the IEEE/CVF Conference on Computer Vision and Pattern Recognition}, pages 2935--2944, 2023.

\bibitem[Yu et~al.(2022)Yu, Wang, Vasudevan, Yeung, Seyedhosseini, and Wu]{yu2022coca}
Jiahui Yu, Zirui Wang, Vijay Vasudevan, Legg Yeung, Mojtaba Seyedhosseini, and Yonghui Wu.
\newblock Coca: Contrastive captioners are image-text foundation models.
\newblock \emph{arXiv preprint arXiv:2205.01917}, 2022.

\bibitem[Yu et~al.(2023)Yu, Shi, Pasunuru, Muller, Golovneva, Wang, Babu, Tang, Karrer, Sheynin, Ross, Polyak, Howes, Sharma, Xu, Tamoyan, Ashual, Singer, Li, Zhang, James, Ghosh, Taigman, Fazel-Zarandi, Celikyilmaz, Zettlemoyer, and Aghajanyan]{yu2023scalingautoregressivemultimodalmodels}
Lili Yu, Bowen Shi, Ramakanth Pasunuru, Benjamin Muller, Olga Golovneva, Tianlu Wang, Arun Babu, Binh Tang, Brian Karrer, Shelly Sheynin, Candace Ross, Adam Polyak, Russell Howes, Vasu Sharma, Puxin Xu, Hovhannes Tamoyan, Oron Ashual, Uriel Singer, Shang-Wen Li, Susan Zhang, Richard James, Gargi Ghosh, Yaniv Taigman, Maryam Fazel-Zarandi, Asli Celikyilmaz, Luke Zettlemoyer, and Armen Aghajanyan.
\newblock Scaling autoregressive multi-modal models: Pretraining and instruction tuning, 2023.

\bibitem[Zhai et~al.(2022)Zhai, Wang, Mustafa, Steiner, Keysers, Kolesnikov, and Beyer]{lit_beyer}
Xiaohua Zhai, Xiao Wang, Basil Mustafa, Andreas Steiner, Daniel Keysers, Alexander Kolesnikov, and Lucas Beyer.
\newblock Lit: Zero-shot transfer with locked-image text tuning.
\newblock In \emph{Proceedings - 2022 IEEE/CVF Conference on Computer Vision and Pattern Recognition, CVPR 2022}, pages 18102--18112, United States, 2022. IEEE Computer Society.
\newblock Publisher Copyright: {\textcopyright} 2022 IEEE.; 2022 IEEE/CVF Conference on Computer Vision and Pattern Recognition, CVPR 2022 ; Conference date: 19-06-2022 Through 24-06-2022.

\bibitem[Zhai et~al.(2023)Zhai, Mustafa, Kolesnikov, and Beyer]{zhai2023sigmoidlosslanguageimage}
Xiaohua Zhai, Basil Mustafa, Alexander Kolesnikov, and Lucas Beyer.
\newblock Sigmoid loss for language image pre-training, 2023.

\bibitem[Zhou et~al.(2017)Zhou, Zhao, Puig, Fidler, Barriuso, and Torralba]{zhou2017scene}
Bolei Zhou, Hang Zhao, Xavier Puig, Sanja Fidler, Adela Barriuso, and Antonio Torralba.
\newblock Scene parsing through ade20k dataset.
\newblock In \emph{Proceedings of the IEEE conference on computer vision and pattern recognition}, pages 633--641, 2017.

\end{thebibliography}
